\documentclass[11pt, a4paper, singlecolumn]{article}

\usepackage{color}

\usepackage[utf8]{inputenc}
\usepackage[english]{babel}

\usepackage[top=2cm, bottom=2cm, left=1.9cm, right=1.9cm]{geometry}

\usepackage{titlesec}
\titlespacing{\section}{0pt}{\parskip}{\parskip}
\titlespacing{\subsection}{0pt}{\parskip}{\parskip}
\titlespacing{\subsubsection}{0pt}{\parskip}{\parskip}

\usepackage{indentfirst}
\usepackage[export]{adjustbox}
\usepackage{graphicx}
\usepackage{amssymb}
\usepackage{amsmath}

\usepackage[dvipsnames]{xcolor}
\usepackage[figurename=Fig.]{caption}
\usepackage{caption}
\usepackage[colorlinks, citecolor=ForestGreen]{hyperref}

\usepackage{cite}

\usepackage{abstract}
\usepackage{multirow}
\usepackage{array}
\newcolumntype{*}{>{\global\let\currentrowstyle\relax}}
\newcolumntype{^}{>{\currentrowstyle}}

\usepackage{subcaption}

\title{\textbf{AUTSL: A Large Scale Multi-modal Turkish Sign Language Dataset and Baseline Methods}}
\date{\vspace*{2pt}}
\author{
	\normalsize \textbf{Ozge Mercanoglu Sincan}\\
	\normalsize Ankara University\\
	\normalsize Computer Engineering Department\\
	\normalsize omercanoglu@ankara.edu.tr
	\and
	\normalsize \textbf{Hacer Yalim Keles}\\
	\normalsize Ankara University\\
	\normalsize Computer Engineering Department\\
	\normalsize hkeles@ankara.edu.tr
}
\begin{document}
\maketitle
%%---------------------------------
%% edit by Hacer Yalim Keles
%\thispagestyle{fancy}
%%---------------------------------

\begin{abstract}{
	\vspace*{-1.5em}
	\it 
	Sign language recognition is a challenging problem where signs are identified by simultaneous local and global articulations of multiple sources, i.e. hand shape and orientation, hand movements, body posture, and facial expressions. Solving this problem computationally for a large vocabulary of signs in real life settings is still a challenge, even with the state-of-the-art models.  In this study, we present a new large-scale multi-modal Turkish Sign Language dataset (AUTSL) with a benchmark and provide baseline models for performance evaluations. Our dataset consists of 226 signs performed by 43 different signers and 38,336 isolated sign video samples in total. Samples contain a wide variety of backgrounds recorded in indoor and outdoor environments. Moreover, spatial positions and the postures of signers also vary in the recordings. Each sample is recorded with Microsoft Kinect v2 and contains color image (RGB), depth, and skeleton  modalities. We prepared benchmark training and test sets for user independent assessments of the models. We trained several deep learning based models and provide empirical evaluations using the benchmark; we used Convolutional Neural Networks (CNNs) to extract features, unidirectional and bidirectional Long Short-Term Memory (LSTM) models to characterize temporal information. We also incorporated feature pooling modules and temporal attention to our models to improve the performances. We evaluated our baseline models on AUTSL and Montalbano datasets. Our models achieved competitive results with the state-of-the-art methods on Montalbano dataset, i.e. 96.11\% accuracy. In AUTSL random train-test splits, our models performed up to 95.95\% accuracy. In the proposed user-independent benchmark dataset our best baseline model achieved 62.02\% accuracy. The gaps in the performances of the same baseline models show the challenges inherent in our benchmark dataset. AUTSL benchmark dataset is publicly available at \href{https://cvml.ankara.edu.tr}{https://cvml.ankara.edu.tr}.
	
}\end{abstract}


\section{Introduction}

Sign language is a visual language that is performed with hand gestures, facial expressions, and body posture. It is used by deaf and speech-impaired people in communication. Since most of hearing people do not know sign language, there is a need to map signs to their associated meanings with computer vision based methods to help communication of the deaf-mute people with the rest of the community. 

Recognition of signs using computational models is a challenging problem for a number of reasons. First, it requires fine-grained analysis of the local and global motion of multiple body parts, i.e. hand, arms, and face. For some pairs of signs, hand gestures look very similar, yet the differences in the facial expressions identify the meaning. In some cases, a very similar hand gesture can impose a different meaning depending on the number of repetitions. Another challenge is the variations of how a sign is performed by different signers, i.e. body and pose variations, duration variance of different parts of the signs etc. Also, variation in the illumination and background makes the problem harder, which is inherently problematic in computer vision. These problems becomes more challenging when the corpus of the signs increases. 

In the literature, the Sign Language Recognition (SLR) research is carried out in two different branches: The first one is isolated SLR \cite{sincan2019isolated, huang2018attention, li2020word, pigou2018beyond, neverova2015moddrop} where a given spatio-temporal sequence is mapped to a sign; the second one is continuous SLR \cite{huang2018video, cui2019deep, guo2018hierarchical, pu2019iterative,koller2018deep} where it is mapped to a sequence of signs. Isolated SLR can be considered as a special kind of action recognition problem.  However, since the hands and face usually cover a small region in video frames, the accurate recognition of a sign imposes different challenges; relatively smaller regions need to be attended accurately. In this research we are focusing on isolated recognition of Turkish Sign Language (TSL) with a large corpus of signs with various challenges. 

Although SLR is an active research area, there is a lack of realistic large-scale sign language datasets. Therefore, most studies are trained and evaluated on either private or public small-scale datasets in the literature \cite{ronchetti2016lsa64, martinez2002purdue, kapuscinski2015recognition, zahedi2005combination, lim2019isolated, cooper2012sign, zafrulla2011american, yang2010chinese, memics2013kinect, unutmaz2019turkish}. However, in order to train a deep learning based sign language recognition model, the amount of training data is crucial. In recent years, larger datasets have been published \cite{huang2018attention, li2020word, chai2015devisign}, which contain a large vocabulary size \cite{chai2015devisign}, large number of samples \cite{li2020word}, with many signers \cite{huang2018attention}. These datasets help building practical SLR models. Although each of them has several challenges, video samples usually have a plain background. This makes it difficult to develop models that can be used in daily life. In the field of TSL, some early domain specific research are conducted for special purposes, e.g., \cite{aran2009signtutor, uluer2015new, alaydin2018query} aims to assist TSL education, \cite{camgoz2016bosphorussign} implements human computer interaction systems in health and finance domains. Due to the absence of publicly available large-scale TSL datasets, researchers have to create their own small scale datasets for the development of special purpose SLR systems \cite{camgoz2016bosphorussign, memics2013kinect, unutmaz2019turkish}. There is a need for a new publicly accessible large-scale TSL dataset to provide the ground for various researches in this domain, especially using the recent deep learning techniques.

In this study, we present a large-scale isolated Turkish Sign Language dataset with Kinect version 2.0 that provides RGB, depth, and skeleton data. It consists of 226 signs performed by 43 different signers and 38,336 isolated video samples. Our dataset differentiates from other publicly available large-scale datasets in that it has 20 different backgrounds with different challenges.  We have focused on user-independent recognition of signs, which we believe is crucial for a model to be useful in practice. Therefore, we provide a benchmark that provides training and test video sets with separate signers in them; the signers in the test set do not appear in the training set. We think that our dataset will both contribute to the progress of studies in the field of TSL and can be a benchmark in general in the isolated SLR domain due to the challenges it provides. 

In this paper, we evaluate our dataset with several deep learning based models that is configured to work with RGB and RGB+Depth (RGB-D) data without any explicit segmentation. The models are trained primarily in CNN + LSTM structure, where features are extracted from each frame separately using a 2D-CNN model and temporal relations of the frames are captured by an LSTM model. In addition to the basic model, different sub-models are integrated in between CNN and LSTM to encode the extracted features in multiple scales and to identify spatio-temporal regions of attention. For this purpose, we plugged in a feature pooling model (FPM) after the CNN model to obtain multi-scale representation of the features using single scale input; and we integrated spatio-temporal attention to the LSTM features. We then generated another model by replacing the LSTM with a bidirectional LSTM (BLSTM) model in the best performing model alternative. We provide empirical results of each model alternative using provided test data, using RGB and RGB-D modalities. 

The rest of this paper is organized as follows. We examine existing SLR datasets and related works in Section \ref{sec:RelatedWorks}. We then introduce our new AUTSL (Ankara University Turkish Sign Language) dataset in Section \ref{sec:AUTSLDataset}. We give the details of our baseline models in Section \ref{sec:methods}. Then, we provide our empirical evaluations in Section \ref{sec:experimentsResults} and conclude the paper in Section \ref{sec:conclusion}.

\section{Related Works}
\label{sec:RelatedWorks}
Similar to many pattern recognition systems, sign recognition systems are composed of two primary components: (1) feature extraction, (2) classification. Extracting the best feature representation of the signs from video streams is a crucial step to obtain higher classification accuracies. Therefore, some previous works explicitly segment hands or/and face before extracting the features; they use colored gloves \cite{grobel1997isolated, lamberti2011real, han2009modelling, ronchetti2016lsa64, cooper2012sign, zafrulla2011american} or data gloves \cite{shukor2015new} to track movements of the hands and deal with segmentation and occlusion problems more accurately. However, the requirement of wearing gloves at all times is not practical in daily life and data gloves with probes often limit the natural movements of the signers. Some other works propose segmenting hand regions by the help of hand motion speed and trajectory information \cite{han2009modelling, cooper2012sign} or skin color detection \cite{dardas2011real, yang2010chinese}. Skin color detection is one of the most popular segmentation method. However, it is sensitive to illumination changes. Also, face and hands could be confused easily with skin-like objects in the background. With the emergence of Microsoft Kinect technology, new modalities such as depth and skeleton are also provided with the RGB data. Some studies utilize depth data for accurate segmentation of the hands \cite{suarez2012hand, ren2013robust}. Depth data is more robust to illumination changes and cluttered background compared to RGB data. It works well to track a large object, e.g., the human body. Skeleton data provides some of the body key-points at the junctions, e.g., neck, elbow, wrist etc. However, it does not cover the details in the fingers of hands, which is crucial for discriminating local hand gestures. Therefore, it is still difficult to segment the human hand with complex articulations even with the different modalities provided with Kinect \cite{ren2013robust}.

Early studies utilized handcrafted features, such as scale invariant feature transform (SIFT) \cite{dardas2011real, yang2010chinese}, histogram of gradient (HOG) \cite{han2009modelling, cooper2012sign, jangyodsuk2014sign}. After feature extraction, features are fed into a classifier such as support vector machine (SVM) \cite{dardas2011real, yang2010chinese}, K-nearest neighbour (K-NN) \cite{memics2013kinect}, or sequence models such as Hidden Markov Models (HMMs) \cite{grobel1997isolated, ronchetti2016lsa64, cooper2012sign, zafrulla2011american}. Also, some studies use dynamic time warping (DTW), a time series matching algorithm, for recognition \cite{han2009modelling, jangyodsuk2014sign}. 

In parallel to the success of the deep learning based models in other domains, many works in the SLR domain recently conduct research using deep neural networks. In these approaches, instead of hand-crafted feature extraction, Convolutional Neural Networks (CNNs) are utilized effectively \cite{sincan2019isolated, li2020word, pigou2018beyond, li2018deep, lim2019isolated, koller2018deep, tur2019isolated, shi2018american, joze2018ms}. While some of these studies do not require any segmentation methods \cite{sincan2019isolated, li2020word, pigou2018beyond, tur2019isolated}, some studies prefer to use neural networks, such as Fast R-CNN and Faster R-CNN, in order to locate the hand region \cite{li2018deep, lim2019isolated, shi2018american}. Recently, attention based models have been successfully applied in other computer vision tasks, such as image captioning \cite{xu2015show} and action recognition \cite{sharma2015action}. These models learn the relevant spatial or temporal parts of the image or video automatically from data. These models have also been used in the SLR domain \cite{shi2019fingerspelling, huang2018attention, guo2018hierarchical, li2018deep, shi2018american}.

The sign language recognition literature is vast and a detailed review of all the literature is outside the scope of our paper. A recent detailed review of SLR works is provided in \cite{SLRReview2019}. In this section, we first overview the existing publicly available large-scale isolated sign language datasets. Then, we review deep learning based sign recognition language methods and attention based models.

\subsection{Sign Language Datasets}
\label{sec:SRLDatasets}

In the literature, most sign language datasets are small-scale in terms of number of signs, number of signers and total sample size, e.g., LSA64 \cite{ronchetti2016lsa64}, Purdue RVL-SLLL \cite{martinez2002purdue}, PSL \cite{kapuscinski2015recognition}, RWTH BOSTON50 \cite{zahedi2005combination}. LSA \cite{ronchetti2016lsa64} is an Argentinian  Sign Language dataset that contains 64 signs that are performed by 10 signers. There are 3,200 RGB samples in total. The signers wore different colored gloves for each hand during recording. Purdue RVL-SLLL \cite{martinez2002purdue} is an American Sign Language dataset that consists of motions, handshapes, signs and sentences performed by 14 signers. It contains 2,576 RGB videos in total. PSL Kinect 30  \cite{kapuscinski2015recognition} and  PSL ToF 84  \cite{kapuscinski2015recognition} are Polish Sign Language datasets that consist of 30 and 84 signs, and in total 300 and 1680 samples, respectively. Both datasets provide RGB and depth modalities. RWTH BOSTON50 \cite{zahedi2005combination} is an American Sign Language that contains 50 signs that are performed by 3 signers. It provides only 483 RGB samples in total. An extended list of sign language datasets can be found in \cite{li2020word, bragg2019sign}.  Montalbano Italian gesture dataset \cite{escalera2014chalearn}, which has recently become one of the most widely used isolated SLR datasets, contains 20 gestures and approximately 14,000 samples in total. It contains 27 signers with variations in background, clothing and lighting. It was recorded with Microsoft Kinect v2 that provides RGB, depth, user segmentation, and skeleton modalities.

%\paragraph
In recent years, a number of large-scale datasets have been published. Table \ref{table1} provides an overview of the large-scale isolated sign language datasets. ASLLVD \cite{neidle2012challenges} has 2,742 signs in American Sign Language (ASL). Although the dataset has large vocabulary size, it has only 9,794 samples in total (3.6 examples per sign on the average). This dataset aims to serve as the basis for development of sign lookup technology in ASL. The video sequences are collected from four cameras simultaneously; two frontal views, one side view, and one view zoomed in on the face of the signer. DEVISIGN \cite{chai2015devisign} is a Chinese Sign Language dataset that consists of 2,000 signs and 24,000 samples that are performed by 8 signers. The videos are recorded with Microsoft Kinect v1, which provides RGB, depth, and skeleton data, in a lab environment in front of a white wall. MS-ASL dataset \cite{joze2018ms} provides 1000 signs, 222 signers, and 25,513 samples. It is collected from a public video sharing platform, i.e. YouTube. Many of videos are performed by ASL students and teachers. In order to provide a basis for signer independent recognition systems, the signers in train, validation, and test set are distinct. It is worth to mention that some of the video links have expired and inaccessible in this dataset \cite{li2020transferring}. CSL \cite{huang2018attention} is a Chinese Sign Language dataset that consists of 500 signs performed by 50 different signers and 125,000 samples. It is recorded with Microsoft Kinect v2 that provides RGB, depth, and skeleton data. Besides being large-scale, this dataset also focusses on user-independent recognition of signs. They select different signers for the training and test sets. The videos are recorded in front of a white background. WLASL \cite{li2020word} is another ASL dataset that consists of 2,000 signs performed by 119 signers and 21,083 samples. Each sign is performed by at least 3 different signers. The dataset consists of only RGB videos. It is collected from 20 different educational sign language websites that provide lookup functions for ASL signs and from ASL tutorial videos on YouTube. In the videos, signers are in a nearly-frontal view with plain background, generally wearing a black colored clothes. We noticed recently in \cite{ozdemir2020bosphorussign22k} that the authors also aim to provide a large-scale TSL dataset, with 744 signs, 6 signers, and 22,542 samples. Since the dataset is not released yet, we preferred not to include it in Table \ref{table1}.

Our AUTSL dataset is a new large-scale Turkish Sign Language dataset with 226 signs, 38,336 samples in total. It is performed by 43 different signers. The average number of samples per sign in our dataset is 169.6, which is the second largest number of samples per sign after the CSL dataset \cite{huang2018attention}. Our dataset differentiates from all aforementioned large-scale datasets in that it has 20 different backgrounds with many challenges, i.e. variation in the lighting, different indoor and outdoor background objects etc. Some of our videos have dynamic backgrounds; some videos that are recorded in the outdoor environments have background objects that move with the wind, and in some recordings, people are passing by behind the signers in the background. In this sense, the samples are collected to provide realistic scenarios for daily use-cases. The details of our AUTSL dataset is given in Section \ref{sec:AUTSLDataset}.
 
\begin{table}[t]
	\caption{Overview of existing large-scale isolated sign language recognition datasets.}
	\centering
	\label{table1}
		\begin{tabular}{ lccp{2cm}lcp{1.7cm} }
			\hline
			\textbf{Datasets} & \textbf{Year} & \textbf{Sign Language}  & \textbf{\centering \#Avg sample per sign} & \textbf{\#Signs}  & \textbf{\#Signers}  & \textbf{\#Total samples} \\ \hline
			ASLLVD \cite{neidle2012challenges}      & 2012 & American & 3.6 & 2,742 & 6 & 9,794   \\ 
			DEVISIGN \cite{chai2015devisign}    & 2014 & Chinese & 12 & 2,000 & 8 & 24,000   \\ 
			MS-ASL \cite{joze2018ms}   & 2019 & American & 25.5 & 1,000 & 222 & 25,513  \\ 
			CSL \cite{huang2018attention}     	& 2019 & Chinese & 250 & 500 & 50 & 125,000 \\ 
			WLASL \cite{li2020word}       & 2020 & American & 10.5 & 2,000 & 119 & 21,083   \\ 
			AUTSL (Ours) & 2020 & Turkish &  169.6 & 226 & 43 & 38,336  \\ \hline
			
		\end{tabular}
\end{table}

\subsection{Deep Learning based SLR Approaches}
\label{sec:SRLApproaches}

In recent years, most studies have been proposed with deep learning based methods. In  ChaLearn 2014 Looking at People Challenge gesture recognition track \cite{escalera2014chalearn}, the winner of the competition \cite{neverova2014multi} proposed a deep neural network, which outperforms other traditional methods. 

In deep learning based methods, basic approach for feature extraction is using CNNs. After feature extraction, while some studies use fully connected layers \cite{neverova2015moddrop, unutmaz2019turkish}, most studies use recurrent neural networks \cite{sincan2019isolated, pigou2018beyond, tur2019isolated, bahdanau2014neural} on top of the CNN models. \cite{neverova2015moddrop} use combination of video in multiple modalities (RGB, depth, intensity), articulated pose and audio streams as inputs. After feature extraction with CNNs, they fuse streams with a set of fully connected layers. They observe that fusing multiple modalities at multiple-scales leads to a significant increase in recognition rates. In \cite{pigou2018beyond}, researchers compare the models that contain CNN architectures, temporal pooling, bidirectional LSTM, or temporal convolutions. They observe that incorporating temporal convolutions and bidirectional LSTM outperforms single-frame and temporal pooling architectures. In \cite{tur2019isolated}, Siamese CNN architecture is used to extract features from the RGB and depth data in parallel. Then, two types of recurrent neural network, LSTM and GRU, are experimented with. In our preliminary work \cite{sincan2019isolated}, we used a feature extraction module (FPM), which is designed with parallel convolutions with different dilation rates, with a pretrained CNN network. Then LSTM is used to model the temporal characteristics of the stream. In the recent years, some studies use 3D-CNNs in order to capture spatial-temporal features together \cite{huang2018attention, li2020word, joze2018ms}. In \cite{li2020word}, pose based and visual appearance based approaches are compared. They compare 2D-CNNs with RNNs and 3D-CNNs for visual appearance based baselines. In their work, 3D-CNNs have higher network capacity, hence achieve better results. Moreover, their model is pretrained both with ImageNet \cite{russakovsky2015imagenet} and Kinetics action recognition dataset \cite{carreira2017quo}. %\cite{joze2018ms} also experiment with both 2D-CNN-LSTM and 3D-CNN networks. They report better results with 3D-CNNs. 

Recent studies also incorporate attention mechanisms into their deep networks in many tasks with promising results. In \cite{bahdanau2014neural}, an attention model is integrated to a bidirectional RNN for English-French machine translation. In \cite{xu2015show}, visual attention model is proposed for image caption generation. They incorporate attention mechanism to an LSTM that generates a weight for each spatial location. Attention weights encodes the importance and relevance of a location for producing the next word. In \cite{sharma2015action}, researchers adapt the attention model of \cite{xu2015show} to action recognition problem. They incorporate spatial attention mechanisms into their deep networks to focus on the regions of interest. Since attention mechanisms achieve promising results in action recognition problem, it also attracts the researchers in the SLR domain. In \cite{li2020word}, an attention based 3D-CNN network is proposed for CSL recognition. On the proposed method, they incorporate spatial attention into 3D-CNN to select skeleton joints of hand and the arm; spatial attention map peaks around these regions. They then feed extracted features into a bidirectional LSTM. They also incorporate temporal attention to LSTM in order to highlight significant video clips. 

In \cite{shi2019fingerspelling}, ASL fingerspelling recognition model is proposed with iterative visual attention mechanism for real-life data. Fingerspelling is a part of sign language in which words are signed letter by letter. It is usually used for spelling proper nouns, e.g., names of people. They use 2D-CNNs pretrained on ImageNet for feature extraction and they feed extracted features to LSTM. ASL fingerspelling signs are only one-handed and the attention mechanism enables the model to focus on active hand region. However, high resolution is needed to get sufficient information; therefore, they aim to retain the highest resolution available while zooming in with iterative attention. In \cite{shi2018american}, an attention-based recurrent encoder-decoders are proposed for ASL fingerspelling problem. In the decoding, temporal attention weights are used to focus on the important visual features when producing each output letter. %In \cite{huang2018video} and \cite{guo2018hierarchical}, attention mechanisms are incorporated to their networks for continuous sign language problem.

\section{AUTSL Dataset}
\label{sec:AUTSLDataset}

%\paragraph
In this section, we introduce our large-scale, multi-modal Turkish Sign Language dataset, named shortly as AUTSL \footnote{{https://cvml.ankara.edu.tr/}}. Our motivation is to collect a large dataset with different challenging backgrounds that is suitable for modelling a realistic SLR system with real life scenarios. Main characteristics of our dataset are summarized in Table \ref{tab:table2}. We record our dataset using Microsoft Kinect v2, hence it contains RGB, depth, and skeleton modalities. We apply some clipping and resizing operations to RGB and depth data and provide them with the resolution of 512x512. The skeleton data contains spatial coordinates, i.e. (x, y), of the 25 junction points on the signer body.

\begin{table}[t]
	\caption{The statistics of AUTSL dataset.}
	\centering
	\label{tab:table2}
	\begin{tabular}{ ll }
		\hline
		\textbf{Property} & \textbf{Description} \\ \hline
		Number of signs       & 226   \\ 
		Number of signers     & 43   \\ 
		Total samples       & 38,336  \\ 
		Number of different backgrounds       & 20   \\ 
		Mean sample per sign    & 169.6   \\ 
		Modalities     & RGB, depth, skeleton  \\ 
		RGB and depth resolution    & 512x512  \\ 
		FPS     & 30  \\ \hline
		
	\end{tabular}
\end{table}

Our dataset consists of 226 signs. When choosing our signs, we paid attention to selecting the signs that are used frequently in daily spoken language. Moreover, we considered to keep a balance in the dataset content to increase the variety of the signs with respect to the motion characteristics of hands while keeping similarly performed different signs at the same time. In this process, we worked with a group of TSL instructors. The selected signs cover a wide variety in terms of hand shape and hand movements. In some of the signs, hands hide each other, e.g., “ayakkabi” (shoe), “bal” (honey), or face, e.g., “beklemek” (wait), “uzgun” (unhappy). In some signs, hands move in the direction of depth, e.g., “itmek” (push), “terzi” (tailor). In some signs, the right hand and left hand are in a cross position, e.g., “yardim” (help), “tehlike” (danger). Some of our signs are compound signs formed by making two consecutive signs. Some of these consecutive signs are also included in our dataset as single signs. For example, "hastane" (hospital) sign is formed by making "doktor" (doctor) and "bina" (building) signs consecutively. The signs for hospital and doctor are both included in our dataset. Similarly, "yemek" (eat) and "ocak" (cooker) signs and the compound versions of two, "yemek pisirmek" (cooking) are also included.

\begin{figure}
	\centering
	\includegraphics[width=1\textwidth]{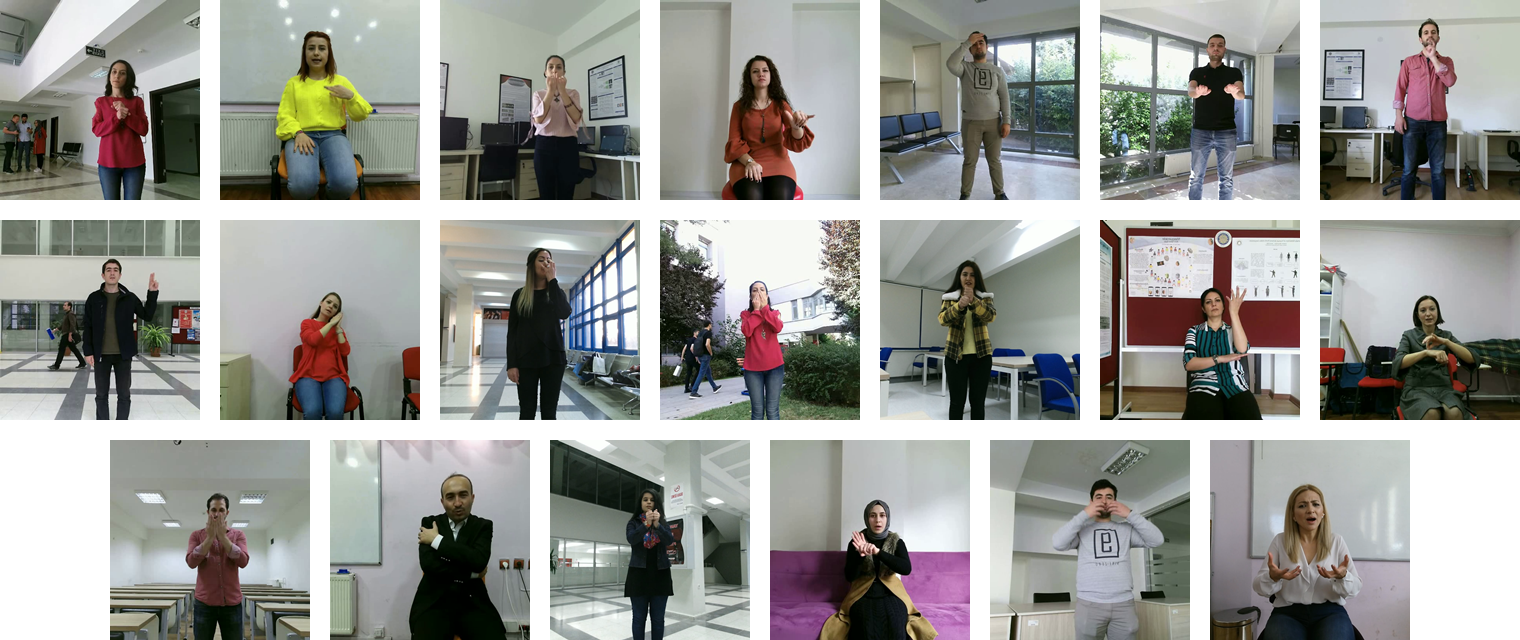}
	\caption{Examples of different backgrounds from AUTSL.}
	\label{fig1}
\end{figure}

We also paid a lot of attention to create AUTSL with various and challenging backgrounds. It contains 20 different backgrounds. For some backgrounds in this set, we also recorded some videos by changing the camera field-of-view, or by adding or removing some objects to/from the background scene to increase the appearance variance more. In Fig. \ref{fig1}, we depict examples of different backgrounds from AUTSL dataset. As shown in the figure, backgrounds contain several challenges; some outdoor recordings contain dynamic backgrounds, i.e. moving trees, or people are passing by behind the signer. Videos contain various lighting conditions, from sunlight to artificial light. Therefore, video frames contain illumination changes and some shadowed or bright-dark areas.

\begin{figure*}
	\centering
	\begin{subfigure}{0.3\textwidth}
		\includegraphics[width=\linewidth]{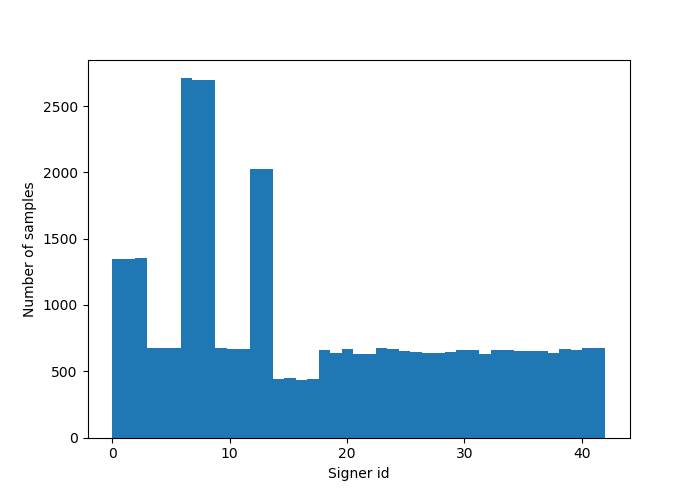}			
		\caption{}
		\label{fig:distrubutions_a} 
	\end{subfigure}		
	\begin{subfigure}{0.3\textwidth}
		\includegraphics[width=\linewidth]{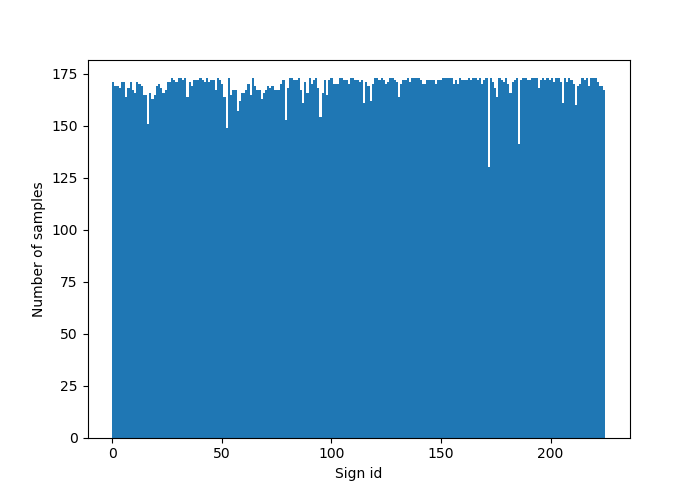}
		\caption{}
		\label{fig:distrubutions_b} 
	\end{subfigure}			
	\caption{Distribution of (a) number of samples performed by each signer and (b) number of samples for each sign.}
	\label{fig2}       % Give a unique label
\end{figure*}

In our dataset, signs are performed by 43 different signers; 6 of them are TSL instructors, 3 are TSL translators, 1 is deaf, 1 is coda (Children of Deaf Adults), 25 are TSL course students and 7 are trained signers who learned the signs in our dataset. 10 of these signers are men and 33 are women; and also, 2 of our signers are left-handed. Fig. \ref{fig2} shows the distribution of the samples over signs and signers. As shown in the figure, we have a balanced dataset according to the sign distribution. On the other hand, the total number of samples for some signers is higher than that of others (Fig. \ref{fig2}b). This is because they are recorded multiple times with different clothes or in different background settings.

\begin{figure}
	\centering
	\includegraphics[width=0.55\textwidth]{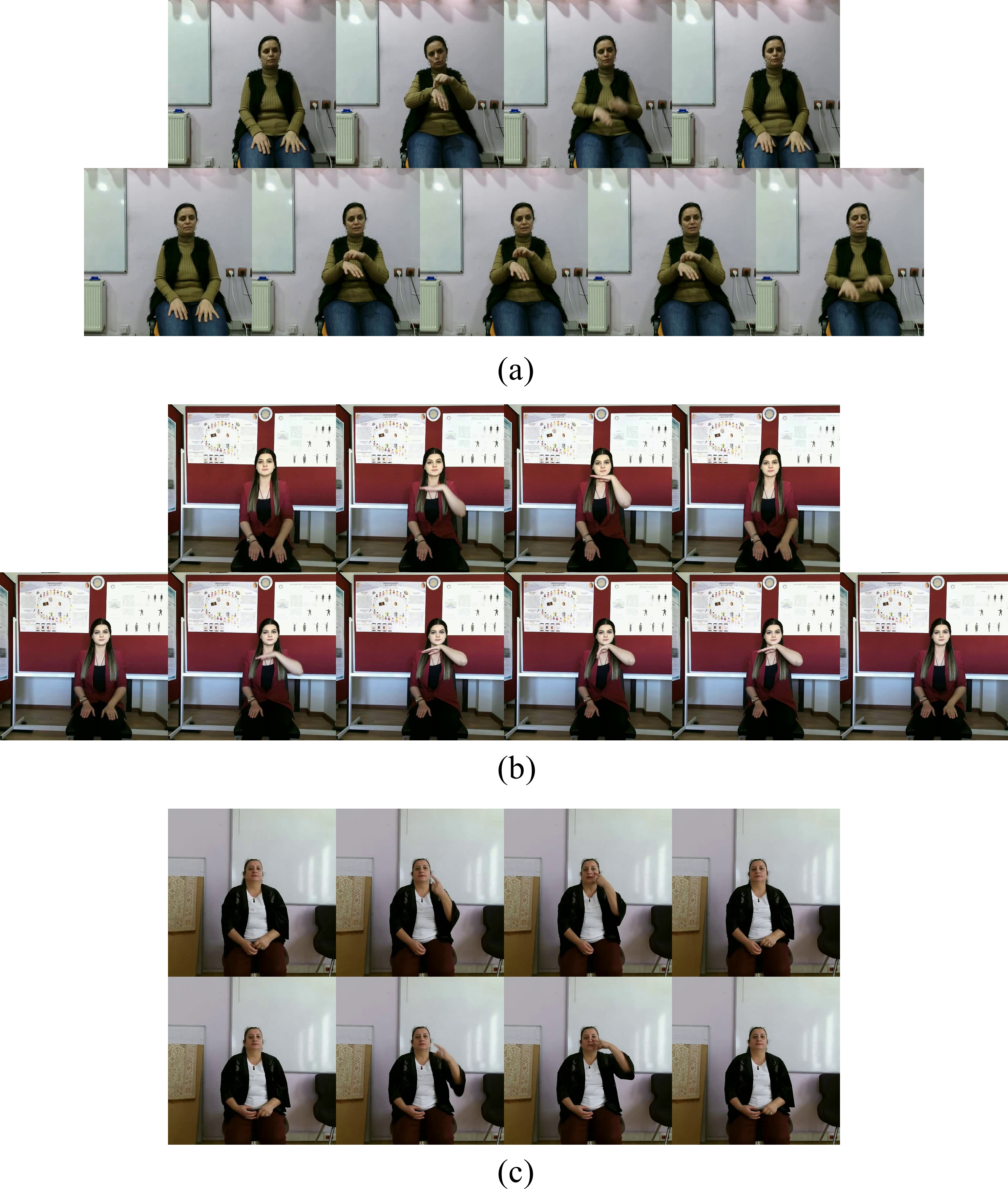}
	\caption{Some of the similar example signs to each other in our dataset. (a) “Doktor” (doctor) and “dakika” (minute) signs differ only in repetition of the hand movement. (b) “Dolu” (full) and “dede” (grandfather) differ only in finger movements. (c) “Devlet” (government) and “mudur” (manager) differ only in the position of the index finger.}
	\label{fig3}
\end{figure}

One of the factors that make our dataset challenging is that it contains very similar signs. For example, as shown in Fig. \ref{fig3}a, although “doktor” (doctor) and “dakika” (minute) signs contain exactly the same hand gesture, they are differentiated according to the repetition cycle of the same gesture. The sign for "doktor" is made by touching the wrist once, while in the sign for "dakika", twice or more. Also, some signs are performed quite similarly in terms of hand shape, hand orientation, hand position or hand movement; changing only one of these factors may mean another sign. For instance, “dolu” (full) and “dede” (grandfather) signs are very similar (Fig. \ref{fig3}b). Although hand shapes, hand rotations and hand positions are very similar, there is only a subtle difference in hand movement. Fingers do not move in the sign of “dolu”, while fingers swing slightly in “dede”. In Fig. \ref{fig3}c, there is only a subtle difference in the position of the hand between “devlet” (government) and “mudur” (manager) signs. In the sign of “mudur”, the index finger touches the nose, and in the “devlet”, it touches under the eye.

In this work, we created a benchmark for user-independent recognition of the signs to observe the performances of the models in a more realistic setting. Therefore, we select 36 signers for training and validation, and the remaining 7 signers for testing. In this setting, our test set contains 9 different backgrounds, 3 of which are not included in the training and validation sets. Our training set contains 27,676 (72\%), validation set contains 4,884 (13\%), and test set contains 5,776 (15\%) samples. In the test set, some signers has relatively more samples than others. Therefore, we will refer to this test set as the imbalanced test set. We also created a balanced test set by making the number of samples of each signer close to each-other by reducing the samples from the signers with excessive samples using random selection. As a result, balanced test set is a subset of the imbalanced test set, which consists of 3,742 samples.

\section{The Methods}
\label{sec:methods}

In order to set a baseline for the evaluation of our AUTSL dataset, we experimented with several deep learning based models. In this section, we first provide the details of the individual components of our models. Following that, we explain our proposed models.

\subsection{Components of the Models}
\label{sec:components_models}
\textit{\textbf{CNN Model: }} Recently, CNNs became the most preferred feature extraction methods in the SLR domain. As we used in our preliminary work \cite{sincan2019isolated, tur2019isolated}, we also selected to use VGG16 model \cite{simonyan2014very} in this work. VGG16 is one of the most used CNN models that is pretrained on ImageNet \cite{russakovsky2015imagenet} dataset to extract features. We use all the convolutional layers of VGG16 model until the last max pooling layer. Since the low-level and mid-level convolutional layers extract generic features, such as edges, corners, common object parts etc., we used the low and mid-level layers as they are without changing the learned parameters. Since high-level layers are more specialized to the objects that are included in the trained dataset, we decided to fine-tune the last two convolutional layers ($conv5\_2$, $conv5\_3$) using our dataset. Before training, we resize the pixel resolutions of the video frames to 256 x 256. When the input images are 256 x 256 x 3, the size of the extracted feature maps at the end of the last convolutional layer become 16 x 16 x 512. 

\textit{\textbf{Feature Pooling Module: }} In \cite{lim2018foreground}, it is shown that using FPM is effective to extract features at multiple scales when single scale input is provided. The idea behind FPM layers is to increase the field-of-views to different sizes in the network using dilated convolutions. We want to assess the performance of FPM in this dataset, considering that multi-scale interpretation of the spatial features may help the network be more aware of the context, i.e. hand, face, body, etc. We showed in our preliminary work \cite{sincan2019isolated} that the FPM module is also effective in isolated sign recognition, using Montalbano dataset. Similar to our preliminary work, we placed the FPM model in this work on top of the last CNN layer. 

\begin{figure}
	\centering
	\includegraphics[width=0.3\textwidth]{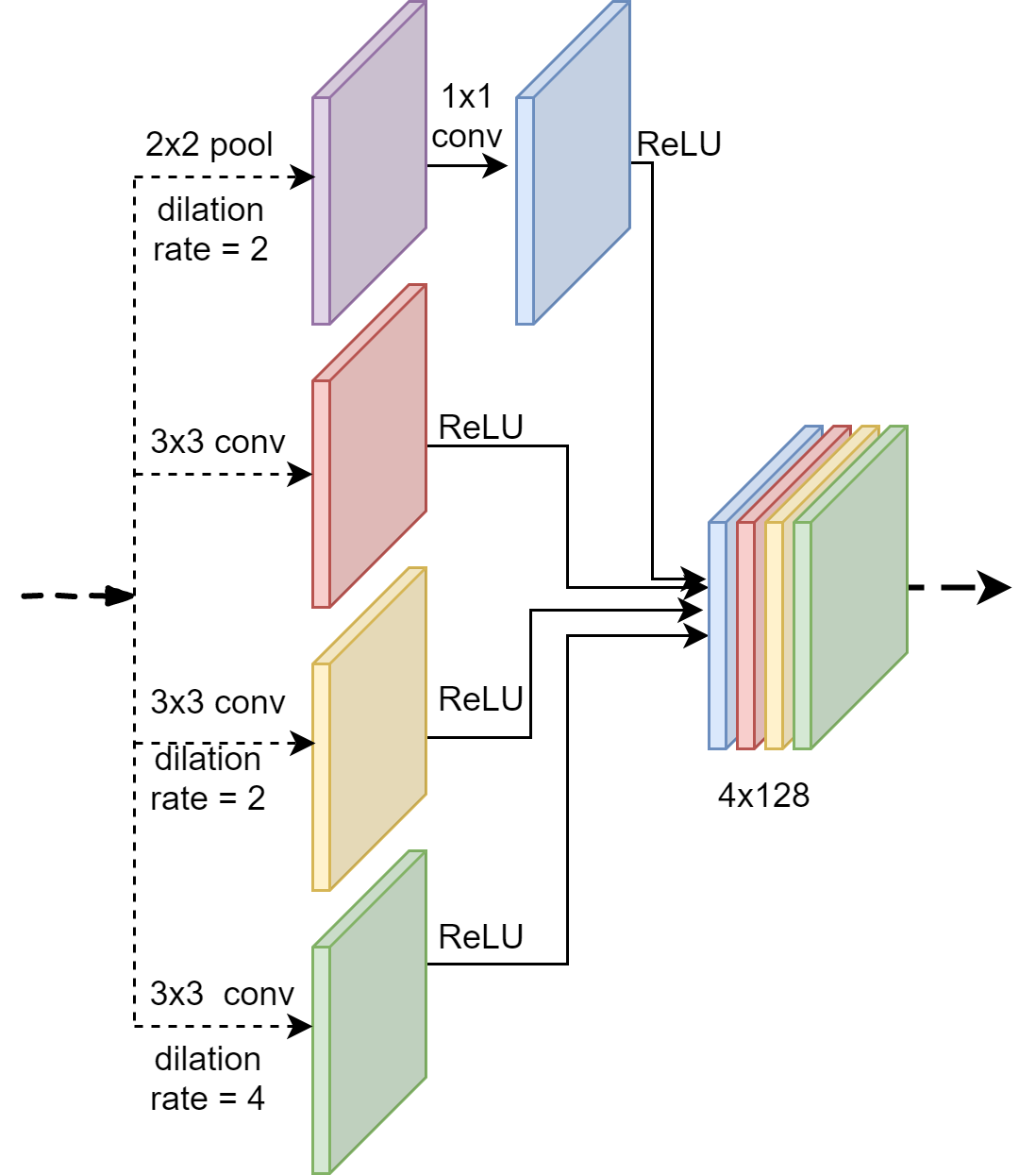}
	\caption{Feature Pooling Module (FPM) \cite{sincan2019isolated}.}
	\label{fig4}
\end{figure}

FPM module is composed of parallel convolutions with different dilation rates. As seen in Figure \ref{fig4}, our FPM module consists of 2x2 max pooling with dilation rate 2 followed by a 1x1 convolution, a normal 3x3 convolution, and two 3x3 dilated convolutions with dilation rates 2 and 4. All the convolutions are implemented with padding, hence the spatial dimensions of the inputs are preserved at the end. The resultant features from the parallel CNN layers are concatenated at the end of FPM. All 4 convolutional layers have 128 output feature planes. Therefore, the resultant shape of the features is 16 x 16 x 512 in our experiments.

\textit{\textbf{LSTM: }} In the literature, recurrent neural networks are commonly used to capture temporal relationship in sequences. In this paper, we use LSTMs \cite{hochreiter1997long} for sequence modelling. After empirically evaluating 1024, 512 and 256 hidden units for LSTMs, we set the number of hidden units to 512 in our architecture, which performed the best with the validation data.  We use random initialization for the hidden and cell states of the first LSTMCell.

\textit{\textbf{Bidirectional LSTM: }} BLSTMs \cite{graves2005framewise} can be considered as extensions to the conventional unidirectional LSTMs, where context of a sequence for each state is coded using the past and the future frames simultaneously. This is achieved using two LSTM models, one for the forward pass, i.e. from the beginning to the end frames; and the other for the backward pass, i.e. from the end to the beginning frames. Hence, each hidden state can aggregate information from the past and the future frames. In our experiments, the $i^{th}$ hidden state is calculated as a concatenation of the corresponding forward and backward hidden states as in \eqref{equ:h_i}:

\begin{equation}
\label{equ:h_i}
h_i= [\overrightarrow{h_i} + \overleftarrow{h_i}]
\end{equation}

We set the number of hidden units to 512 for both forward and backward LSTMs. Therefore, the hidden state sizes of BLSTM become 1024 in our experiments.

\textit{\textbf{Attention Model: }} We integrate a temporal attention mechanism to LSTM and BLSTM models in order to select the most effective video frames in classification. We adapt the temporal attention model proposed by \cite{bahdanau2014neural, raffel2015feed}  to the isolated SLR problem.

In our simple LSTM model, we use the last hidden state, $h_t$, for prediction of a sign. However, in attention-based LSTM, we produce a context vector, $c$, using a weighted sum of all hidden states that are generated for each frame in a video by the LSTM model. This context vector is sent to the fully connected layer for the prediction of the sign. Each hidden state contributes the context vector according to its attention weight. Context vector is calculated as follows:

\begin{equation}
\label{equ:c}
c =\sum_{i=1}^{T}{\alpha}_i h_i
\end{equation}

\begin{equation}
\label{equ:alpha}
{\alpha}_i=\frac{exp(e_i)}{\sum_{k=1}^{T}{exp(e_k)}}
\end{equation}

\begin{equation}
\label{equ:e}
e_i=v^T tan(W h_i + b)
\end{equation}

where ${\alpha}_i$ is the attention weight for the hidden state corresponding to the input frame features, $x_i$. It is calculated by normalizing the attention scores, i.e. $e_i$, with the softmax function as in \eqref{equ:alpha}. Thus, the sum of the weights of all frames is normalized to 1. The higher the score for an input frame, the higher its contribution to the context vector. $e_i$ is produced by a neural network which generates a score for the input features, $x_i$, depending on its hidden state, $h_i$. This neural network is our attention network and it is parametrized by $v, W, b$, where $v {\in} R^d,  W{\in} R^{dxd}, b {\in} R^d$. These parameters are learned during training the models. In this setting, $d$ is the dimension of hidden unit in the LSTM, which is 512 in our experiments.

\subsection{Baseline Models}
\label{sec:baseline_models}

\begin{figure}
	\centering
	\includegraphics[width=1\textwidth]{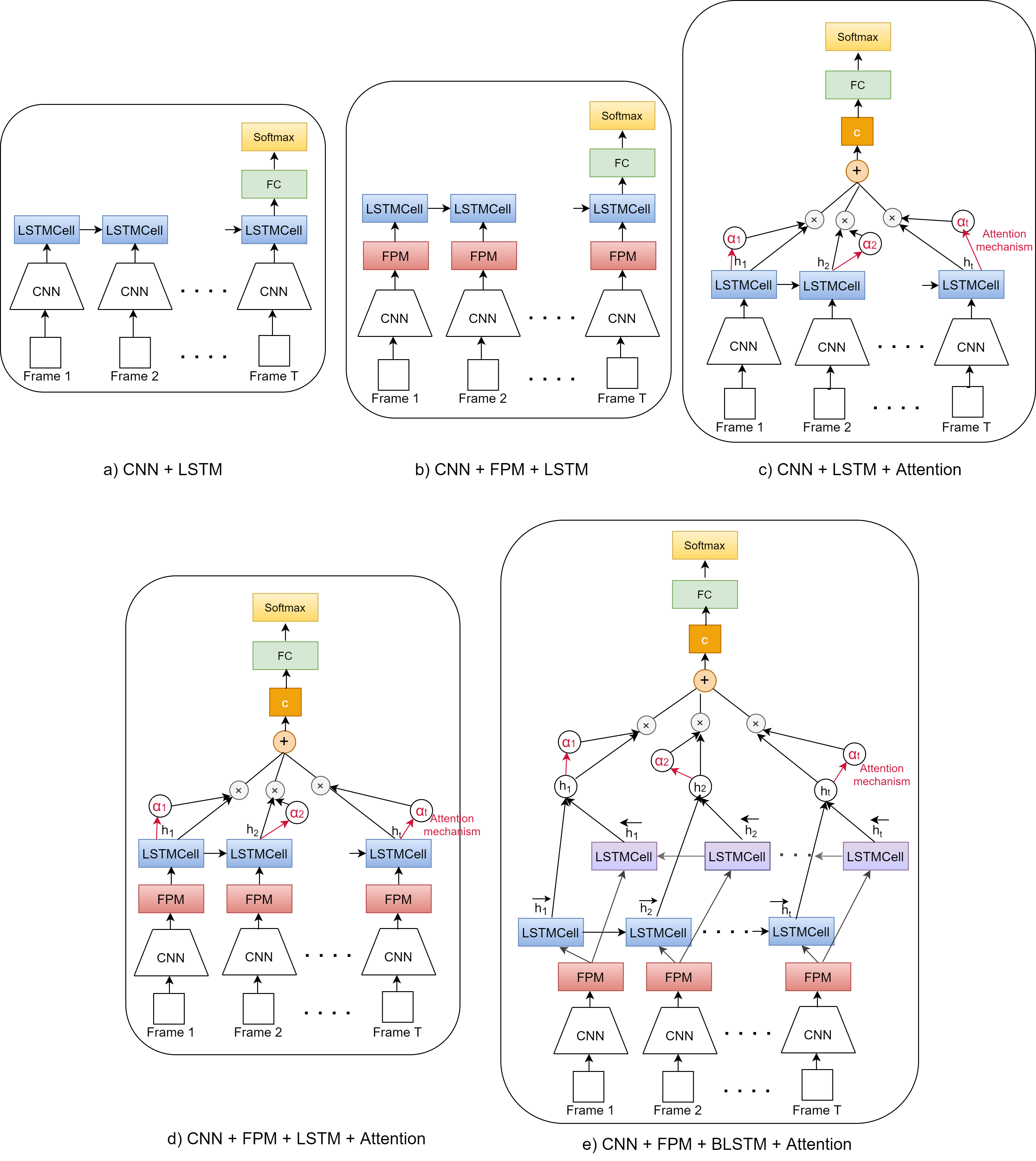}
	\caption{Architectures of our baseline models.}
	\label{fig5}
\end{figure}

We construct five deep neural networks for the empirical evaluations. In all the models, we use CNNs to extract spatial features from each frame. In our experiments, we investigate the contributions of using a feature pooling module and temporal attention model as we described in Section \ref{sec:components_models}. We also compare the performances using simple unidirectional LSTM and bidirectional LSTMs. All our networks, as illustrated in Fig. \ref{fig5}, are separately trained end-to-end.  

\textit{\textbf{CNN + LSTM Model: }} In our models, we conduct our experiments using only RGB and RGB-D modalities, with minor modifications. In order to use the depth data, which is represented as a single channel gray-scale image for each frame, with the pretrained VGG model, we repeat the same depth data into three color channels as in \cite{eitel2015multimodal}. Then, RGB and depth modalities are given as inputs to the two parallel VGG models with exact same architectures and applying similar training regime as we described in the previous section. CNN networks extract features and generate two feature matrices, i.e. one for the RGB data and one for the depth data. Then, we apply global average pooling and reduce the feature map dimensions to a vector of size 512 for each modality, separately. In the RGB only model, we feed 512-dimensional feature vectors into the LSTM model. On the other hand, in the RGB-D network, we concatenate two feature vectors with late fusion and obtain a 1024-dimensional feature vector. LSTM model generates scores using the the last hidden state vector, i.e. $h_t$, after passing it to the Fully Connected (FC) layer. Since we have 226 signs, FC layer is set to have 226 neural units. The scores of the FC layers are fed to a softmax classifier. We refer to this model as \textit{CNN + LSTM} from now on.

\textit{\textbf{CNN + FPM + LSTM Model: }} In the second model, our motivation is to represent the generated features in multiple-scales, so that we can get more contextual clues for classification of individual signs. We add an FPM module after the last CNN layer for that purpose. After that, we apply global average pooling to the extracted features. In the RGB-D model, we again concatenate the two feature vectors with a late fusion. Then, we send extracted features to LSTM. All the architectures, i.e. CNN and LSTM, are the same with the previous model, except for the addition of the FPM module in between these models. As we stated before, all the parameters of this network is trained end-to-end from scratch. We refer to this model as \textit{CNN + FPM + LSTM}.

\textit{\textbf{CNN + LSTM + Attention Model: }} Attention mechanisms have recently shown considerable improvements to many computer vision tasks. Therefore, we also want to investigate the contribution of attention to the classification performance with our dataset. The architecture is designed as follows: First, we extract the features with CNN and then apply global average pooling as in \textit{CNN + LSTM} model. The only difference of this method from \textit{CNN + LSTM} model is that we incorporate a temporal attention mechanism to the features that are passed to the LSTM model. We produce a context vector, c, using all the hidden states as we explained in detail in Section \ref{sec:components_models}. We then send this context vector, instead of the last hidden state, to the FC layer. Finally, we use a softmax classifier. This model is referred to as \textit{CNN + LSTM + Attention}.

\textit{\textbf{CNN + FPM + LSTM + Attention Model: }} In this model, we observe the contribution of using both FPM and a temporal attention mechanism. At first, we extract features with CNN and pass the resultant feature maps to FPM. Then, we use the attention-based LSTM. Here again, we send the context vector to the FC layer and use softmax classifier. This model is referred to as \textit{CNN + FPM + LSTM+ Attention}.

\textit{\textbf{CNN + FPM + BLSTM + Attention Model: }} Finally, we want to investigate the classification performance using bidirectional LSTMs with AUTSL dataset. We configured the components of the model as in the \textit{CNN + FPM + LSTM + Attention} model, but we use attention based BLSTM instead of LSTM this time.

\section{Results and Discussion}
\label{sec:experimentsResults}
We evaluate our baseline models on our new large-scale AUTSL dataset and Montalbano Italian gesture dataset. For AUTSL dataset, our main experiments are configured in a user-independent setting; we use 36 signers for training and validation, and the remaining 7 signers for testing.  We also conducted experiments by randomly selecting the training, validation and test set to evaluate our model performances in user-dependent test setting. In addition to the AUTSL experiments, we also trained our best performing model using the Montalbano dataset. In this section, we first give the evaluation metric. Then, we provide our experimental results.

\subsection{Evaluation Metric}
\label{sec:evaluationMetric}
In order to evaluate the performances of the models, we use the recognition rate, $r$, as defined in  \cite{wan2016chalearn}:
\begin{equation}
	\label{equ:r}
	r=\frac{1}{n} \sum_{i=1}^{n}f(p(i), y(i))
\end{equation}

where $n$ is the total number of samples; $p$ is the predicted label; $y$ is the true label; if $p(i)= y(i), f(p(i),y(i))=1$, otherwise $f(p(i),y(i))=0$.

We will refer to this metric as top-1 recognition rate, since we are only evaluating a model’s best guess. In AUTSL dataset, some of the signs are quite similar to each other; they can be confused by the models. Therefore, in addition to top-1 recognition rate, we also considered top-3 and top-5 recognition performances of the models. Top-N recognition rate refers to the rate by which the true class label exists in a model’s top-N predictions.

\subsection{Experiment Results on AUTSL}

\subsubsection{Training Details}
\label{sec:trainingDetails}
We configured all our model experiments using the same hyperparameters. Since the videos in AUTSL dataset contain variable frame lengths, during training each sample is sent to the network separately; hence we set the batch size to 1. We implemented all the models using PyTorch library \cite{paszke2019pytorch}. In LSTM and BLSTM implementations with variable frame lengths, we use \textit{LSTMCells} units of PyTorch. In order to avoid overfitting, we include dropout layers before sending the features to LSTM/BLSTM models and before the FC layer with dropout rate 0.25. We optimized the multi-class cross-entropy loss using Adam optimizer \cite{konur2015adam}. We set the learning rate to $1e-5$ and reduce the learning rate to $2e-6$, if no improvement is observed in validation accuracies for ten epochs. If there is no improvement for ten of epochs again, we terminate the training process.

\subsubsection{Results}
\label{sec:results}

We conducted a number of experiments to measure the contribution of the use of FPM and the attention model. We also measure the contribution of using multiple modalities, i.e., RGB-D, versus using only RGB. Table \ref{tab:autsl_rgbd} and Table \ref{tab:autsl_rgb} shows the recognition rates of our baseline models using RGB-D and RGB data, respectively.

\begin{table} 
	\caption{Recognition rates (\%) of our models using RGB+Depth data.}
	\centering
	\begin{tabular}{l|ccc|ccc}
		\hline
		\multirow{2}{*}{\textbf{Method}} &
		
		\multicolumn{3}{c}{\textbf{Balanced Test Set}} & \multicolumn{3}{c}{\textbf{Imbalanced Test Set}}                                                                                                                                                                       \\ 
		& {\textbf{top-1}} & {\textbf{top-3}} & {\textbf{top-5}} & {\textbf{top-1}} & {\textbf{top-3}} & {\textbf{top-5}} \\ \hline
		
		CNN + LSTM  & 39.31 & 59.13 & 66.64 & 37.84 & 57.68 & 65.30 \\
		CNN + FPM + LSTM & 41.26 & 60.60 & 68.76 & 39.45 & 58.50 & 66.55 \\ 
		CNN + LSTM + Attention  & 57.80 & 76.24 & 82.57 & 54.55 & 62.79 & 70.82 \\  \hline
		CNN + FPM + LSTM + Attention  & 60.02 & 78.00 & \textbf{83.93} & 56.99 & 75.79 & \textbf{82.22} \\ 
		CNN + FPM + BLSTM + Attention  & \textbf{62.02} & \textbf{78.11} & 83.45 & \textbf{59.24} & \textbf{76.03} & 81.60 \\ \hline		
		
	\end{tabular}
	\label{tab:autsl_rgbd}
\end{table}

\begin{table} 
	\caption{Recognition rates (\%) of our models using only RGB data.}
	\centering
	\begin{tabular}{l|ccc|ccc}
		\hline
		\multirow{2}{*}{\textbf{Method}} &
		
		\multicolumn{3}{c}{\textbf{Balanced Test Set}} & \multicolumn{3}{c}{\textbf{Imbalanced Test Set}}                                                                                                                                                                       \\ 
		& {\textbf{top-1}} & {\textbf{top-3}} & {\textbf{top-5}} & {\textbf{top-1}} & {\textbf{top-3}} & {\textbf{top-5}} \\ \hline
		
		CNN + LSTM  & 23.00 & 37.03 & 43.66 & 22.80 & 36.94 & 43.61 \\ %\hline
		CNN + LSTM + Attention  & 42.14 & 61.83 & 71.21 & 40.89 & 60.68 & 69.42 \\ 
		CNN + FPM + LSTM + Attention   & 44.89 & 64.24 & 72.26 & 43.69 & 62.79 & 70.72 \\
		CNN + FPM + BLSTM + Attention   & \textbf{49.22} & \textbf{68.89} & \textbf{75.78} & \textbf{47.62} & \textbf{67.38} & \textbf{73.89}  \\ \hline		
		
	\end{tabular}
	\label{tab:autsl_rgb}
\end{table}

The challenges inherent in the AUTSL samples are visible in the recognition rates for user independent evaluations. The performance of the vanilla \textit{CNN + LSTM} model using only RGB data is only 23\% with the test data. When we fuse RGB and depth features, the recognition performance significantly increases up to 39.31\%, around 16\% higher than the RGB modality. This is something we expected, since AUTSL contains samples where hands move forward and backward with respect to the camera's optical axis. We think that RGB data alone is not sufficient to accurately discriminate such signs. In this respect, multiple data modalities that we provide with AUTSL is necessary for better identifying some signs. Moreover, as the top-3 and top-5 performances are considered, the recognition rates increase around 20\% and 27\% in RGB-D data with respect to its top-1 accuracy, respectively. These results clearly reveal that the vanilla model confuses some signs; although the true sign is identified 66.64\% of the time in its top-5 predictions, (and 59.13\% of the time in its top-3), the model picks another similar sign in its top-1 order. For imbalanced test, the performance is quite similar; yet slightly worse than the balanced test. Remember that, in imbalanced test set, all the video samples that we have with the selected 7 signers are included. Apparently, the additional samples include more samples of the confused signs; also additional videos containing different backgrounds from the outdoor environment reduce the classification accuracies. For RGB data, top-3 and top-5 predictions are 14\% and 20\% higher than its top-1 predictions, respectively. Although this is a good sign, since it identifies comparatively a good deal of correct signs in its top-3 and top-5 predictions, it is less than RGB-D data with a high margin; in the balanced test, RDB+Depth top-5 predictions are around 23\% higher than RGB top-5 predictions. 

After observing the performance of RGB-D data (Table \ref{tab:autsl_rgbd}), we first completed the experiments by including FPM and attention modalities incrementally using these modalities together. Then after evaluating the performances, we repeated similar experiments using only RGB data. We aim to identify the setting with RGB only data that performs the best. 

We will go over each case separately below:

\textit{\textbf{Results of RGB-D Data:}} We first plug our FPM model to the vanilla \textit{CNN + LSTM} model. FPM improved the recognition rates only slightly, i.e. 1.95\% in the balanced test, 1.61\% in the imbalanced test. The improvement with only FPM model is limited. We then integrated attention to the vanilla model, without FPM first, to see its effect alone to the classification performance. The temporal attention model that we integrated into LSTM model improved the results significantly, i.e. 18.49\% top-1. This improvement reflects to top-3 and top-5 performances as well; top-3 recognition rate of the model becomes 76.24\% and top-5 becomes 82.57\% in the balanced test. The imbalanced test results are also improved in a parallel manner, i.e. 16.71\% in top-1 accuracy. We then plugged in FPM model to the \textit{CNN + LSTM + Attention} model to see its contribution again. It improves the top-1 performances slightly by 2.22\%. After these observations, we set our baseline model for future researches with AUTSL dataset as \textit{CNN + FPM + LSTM + Attention} model. 

\textit{\textbf{Results of RGB Data:}} We conducted similar experiments with RGB only data. Similar to RGB-D models, RGB model top-1 performances increased incrementally in the order we plugged temporal attention model and FPM model, from 23.00\% to 42.14\% and 44.89\%, respectively. Attention model increased the performance significantly here as well, by more than 19\% and FPM improved that performance 2.75\% more. The addition of attention, however, increases the robustness of the predictions with RGB data more than we expected, as far as its top-3 and top-5 predictions are considered. The top-3 prediction of the RGB only model improves 24.80\% more than the vanilla \textit{CNN + LSTM} model. Similarly, top-5 predictions improves 27.55\% more than the vanilla model's top-5 prediction. Still yet, there is quite a margin, i.e. 15.13\%, between the top-1 predictions of RGB+Depth data and RGB only data with the balanced tests using \textit{CNN + FPM + LSTM + Attention} models. The results with the imbalanced tests are also similar. Therefore, depth data provides a significant contribution to the recognition performance with AUTSL dataset.

In addition to using unidirectional LSTM model, we also tested the best model replacing it with a bidirectional LSTM model. The performances are similar, only slightly better in both RGB only and RGB-D modalities. Although the performance of BLSTM model is slightly higher than unidirectional LSTM, we want to underline an issue with BLSTMs that in a real-time application environment, where frames are evaluated online, backward evaluation requires buffering the incoming frames and evaluations can start only after all the frames of an isolated sign is completed. This complicates the process. Additional design issues would emerge while working in continuous sign recognition setting. 

As mentioned earlier in Section \ref{sec:AUTSLDataset}, different signs have very similar gestures in our dataset. In continuous sign recognition, similar signs can be correctly discriminated from the context, the lack of context in isolated recognition makes correct classification harder. Therefore, considering top-1, top-3 and top-5 recognition rates are useful to interpret the performances of the models. As seen in the tables, when comparing the top-1 and the top-3 scores, there is a significant increase in the results. That is, even if a sign cannot be correctly classified in the first order, it can be classified correctly in the first 3 predictions.

\textit{\textbf{About Confused Signs: }} We examine the confusion matrix of our best model, \textit{CNN + FPM + BLSTM + Attention} in the case of fusing RGB and depth data. On the balanced test set, there are around 17 samples for each sign.  We observe that some of the signs are confused more with particular signs. One of the confused sign pairs, the sign “dede” (grandfather) and “dolu”(full), are shown in Fig. \ref{fig3}b. Although there are 17 samples from the sign “dede” in the test set, it is confused 10 times with the sign “dolu”, because these two signs are performed very similarly in hand shape, hand rotation and hand position. We observe that the number of the correct predictions is quite low for some signs. These signs are generally confused with similar sign pairs in the dataset. The increase in the top-3 and top-5 evaluations also reveals this issue. 

\begin{figure}

	\centering
	\begin{subfigure}{1\textwidth}
	\includegraphics[width=1\textwidth]{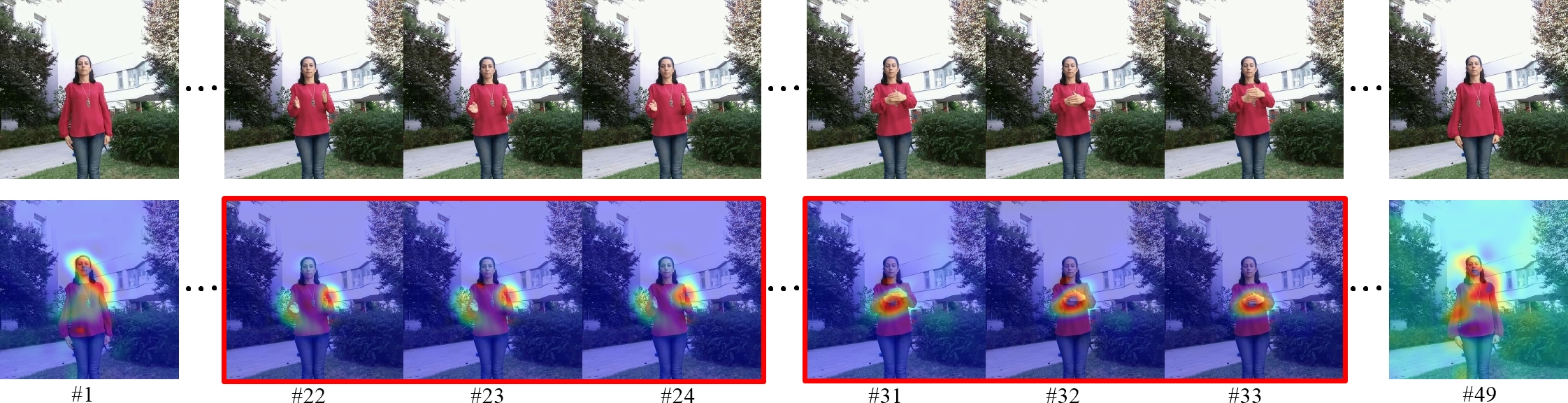}
	\caption{}
	\label{fig:visualize_model_a}
	\end{subfigure}

	\begin{subfigure}{0.6\textwidth}
		\includegraphics[width=1\textwidth]{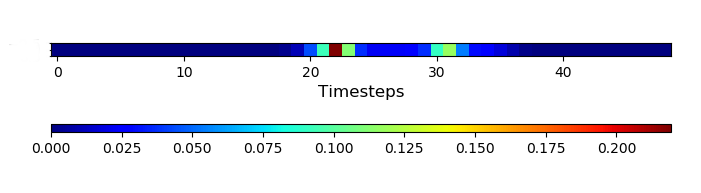}
		\caption{}
		\label{fig:visualize_model_b}
	\end{subfigure}

	\begin{subfigure}{1\textwidth}
	\includegraphics[width=1\textwidth]{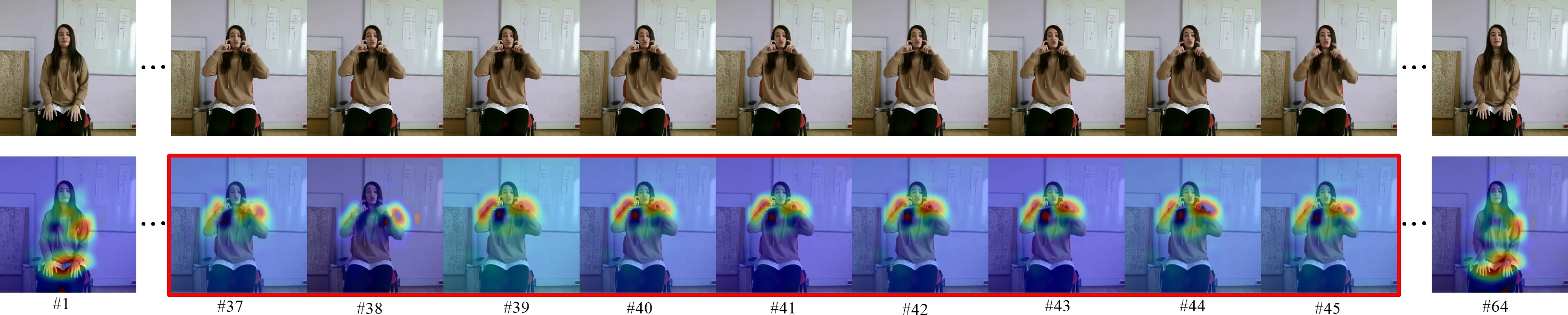}
	\caption{}
	\label{fig:visualize_model_c}
	\end{subfigure}

	\begin{subfigure}{0.6\textwidth}
	\includegraphics[width=1\textwidth]{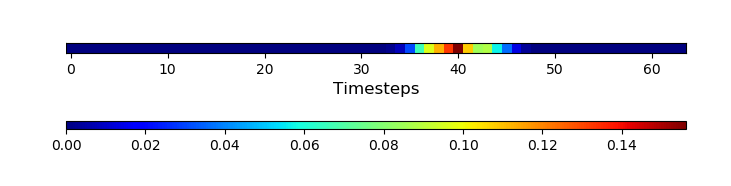}
	\caption{}
	\label{fig:visualize_model_d}	
	\end{subfigure}

\caption{Sample RGB video sequences and GradCAM \cite{selvaraju2017grad} visualizations of the attended regions for two signs: (a) “oda” (room), (c) "fotograf" (photograph). Note that red regions in a frame show highly attended parts, blue regions are attended less. (b, d) Temporal attention weights of the videos. The attended frames are enclosed within a red frame.}
\label{fig6} 
\end{figure}

\begin{figure}
	\centering
	\includegraphics[width=1\textwidth]{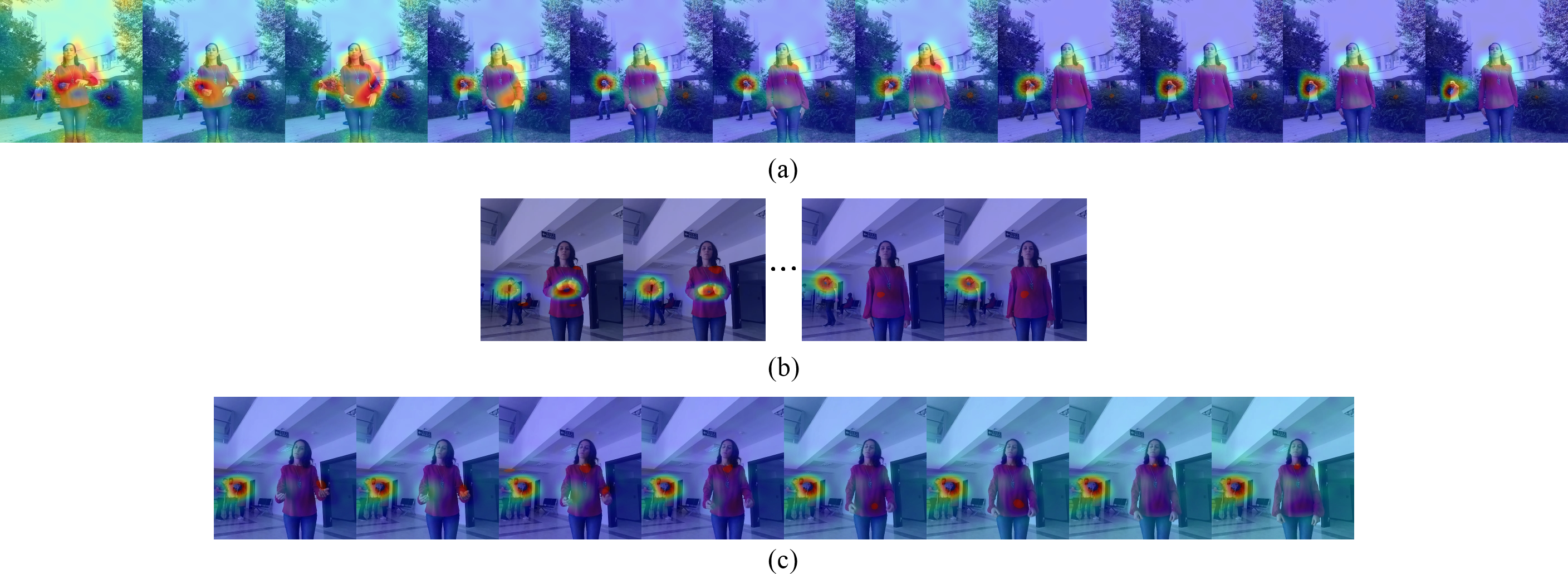}
	\caption{Sample misclassifications due to dynamic backgrounds: (a) "oda" (room), which is the same sign with Fig. 6a. This time, it is misclassified because of the moving person in the background (b) "oda" (room) sign in indoor (c) "gecmis\_olsun" (get\_well).}
	\label{fig7}
\end{figure}

\textit{\textbf{Visualization Results: }} After the quantitative analysis of the proposed models, we also observed the attended spatial regions of the test samples using Grad-CAM \cite{selvaraju2017grad} visualization technique. In addition, we analyzed the distribution of the temporal attention weights over the video frames to interpret the frames that contribute more to the classification (Fig. \ref{fig6}). In the visualizations, we used our \textit{CNN + FPM + BLSTM + Attention} model that is trained using RGB data only. In general, the model learns to focus on the hands, arms, and faces of the signers in the spatial RGB domain. We generated visualizations considering the CNN output layer, before the FPM model. Since FPM model provides multi-scale interpretation of the CNN output features, visualizations generated by CNN outputs look more condensed and sharp on the image domain. Still, when we visualize the spatially attended regions, CNN models that are followed with an FPM model can focus the relevant regions more successfully. 

\begin{figure}
	
	\centering
	\includegraphics[width=0.6\textwidth]{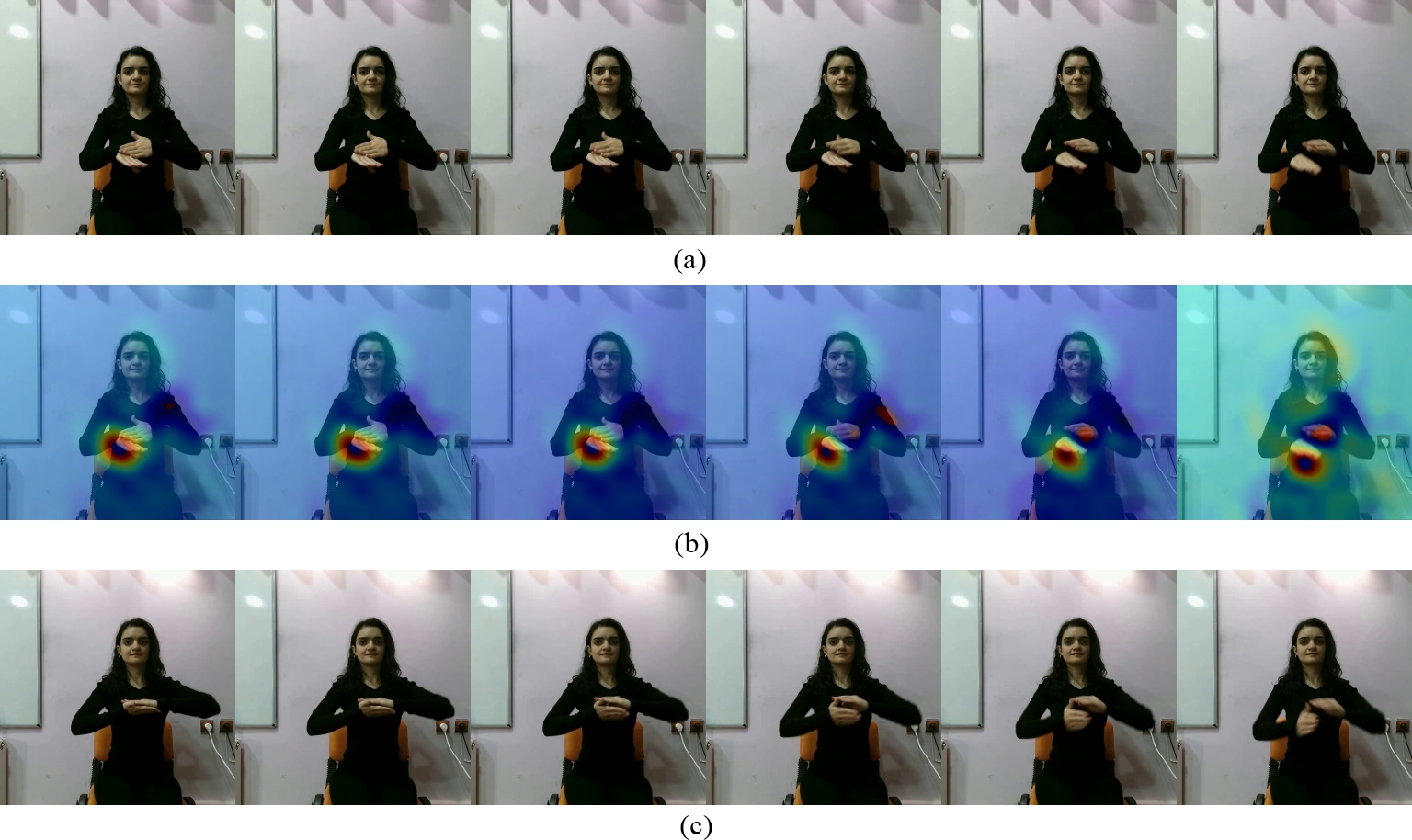}	
	\caption{Sample misclassification due to sign similarity. (a) "yavas" (slow) sign, (b) GradCAM visualization of (a), (c) "arkadas" (friend) sign.}
	\label{fig8} 
\end{figure}

In the time domain, we enclosed the attended frames, which have relatively high attention weight values throughout the whole video, within red a bounding box (Fig. \ref{fig6}). The distributions of the attention weights are visible in Fig. \ref{fig6}b and \ref{fig6}d. As can be seen from the enclosed frames, the attention model highlights the motion sequence that are particularly important for that sign. In other words, it learns to discard the initial and end parts of the video frames. We observed this pattern in almost all the signs. Depending on the particular sign, the weight distribution of the frames are also adapted successfully. This helps the discrimination of signs a lot, since in our dataset signers start performing the sign from a neutral position, i.e. hands are stationary down below, and end similarly, i.e. the hands return back to neutral position. These analysis support the obtained quantitative increase in the classification accuracies when temporal attention is integrated to the models.

Visualizations of the attended regions are also useful to interpret the reason behind our model's poor performance for some signs. We show some samples that are all misclassified due to dynamic background in Fig. \ref{fig7}. In all these three samples, some people are passing by behind the signer, both in indoor and outdoor settings; both spatial attention and temporal attention is badly influenced by the appearance and motion of another person on the scene. Although they appear small in the background, far behind the signer, the spatial attention shifts to those people. The sample sign shown in Fig. \ref{fig7}-middle part is the same sign with Fig. \ref{fig6}a. It is misclassified by our model this time. In addition to the attention shift in the spatial domain, the attention in the temporal domain is also affected by the motion in the background; our model attends to the last two frames this time, where the signer has already settled in neutral position of ending the sign. In that case, there is no spatially interesting motion of our signer to attend; so it focuses on the person in the background and misclassifies the sign that it was classifying correctly in the absence of disruption.

We also provide a sample visualization of the attended regions and frames for a confused sign pair (Fig. \ref{fig8}). The signs corresponding to "yavas" (slow) and "arkadas" (friend) are performed similarly in hand positions and shapes. Although the model pay attention to the hands and semantically relevant frames in time, the sign, which is depicted in the last row of Fig. \ref{fig8}, is misclassified as "arkadas" sign since they look quite similar. In such cases, the correct sign is usually included in the model's top-3 or top-5 predictions.

\begin{figure} 
	\centering
	\begin{subfigure}{0.32\textwidth}
		\includegraphics[width=\linewidth]{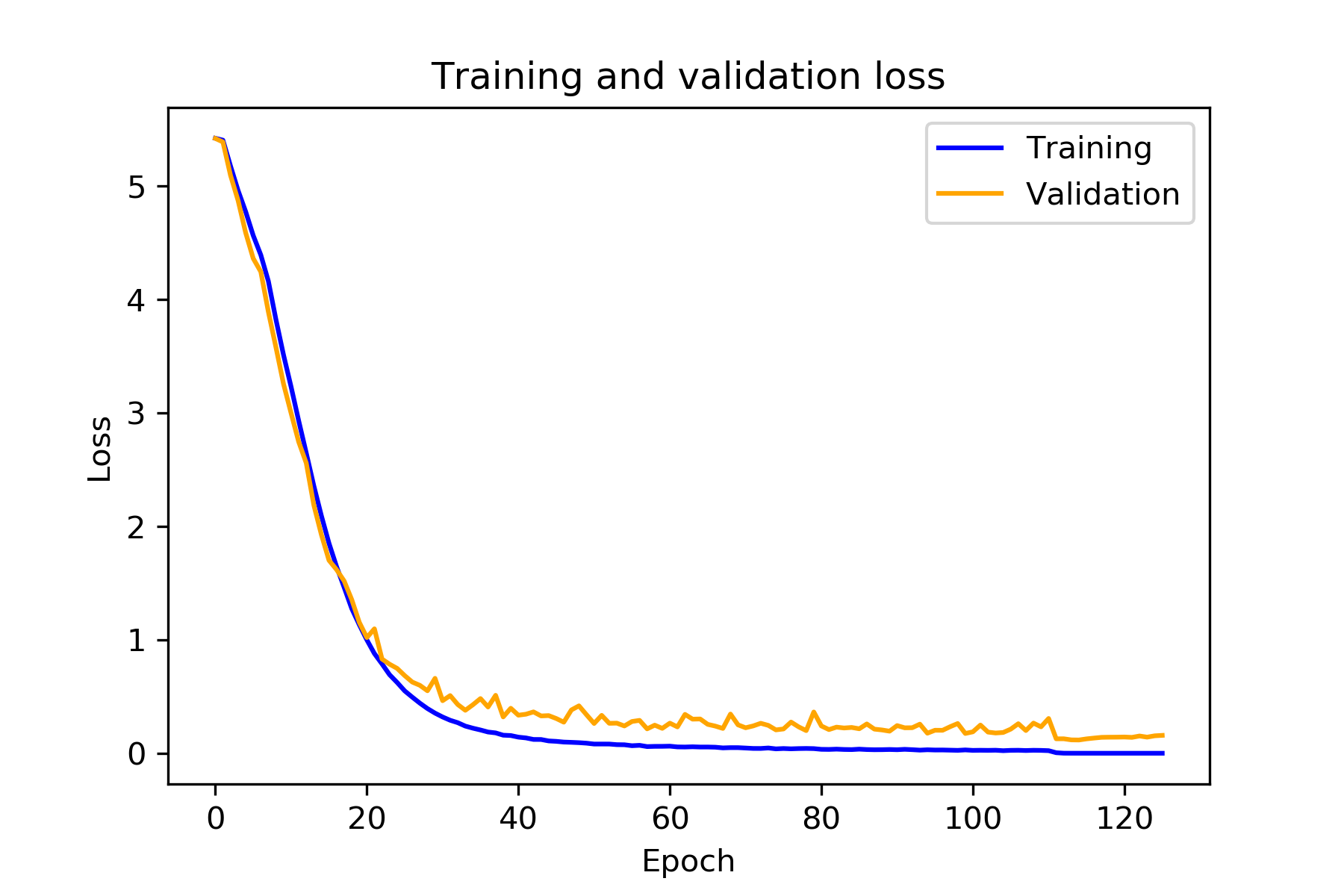}	
		\caption{CNN + LSTM}
		\label{fig:44_3}	
	\end{subfigure}		
	\begin{subfigure}{0.32\textwidth}
		\includegraphics[width=\linewidth]{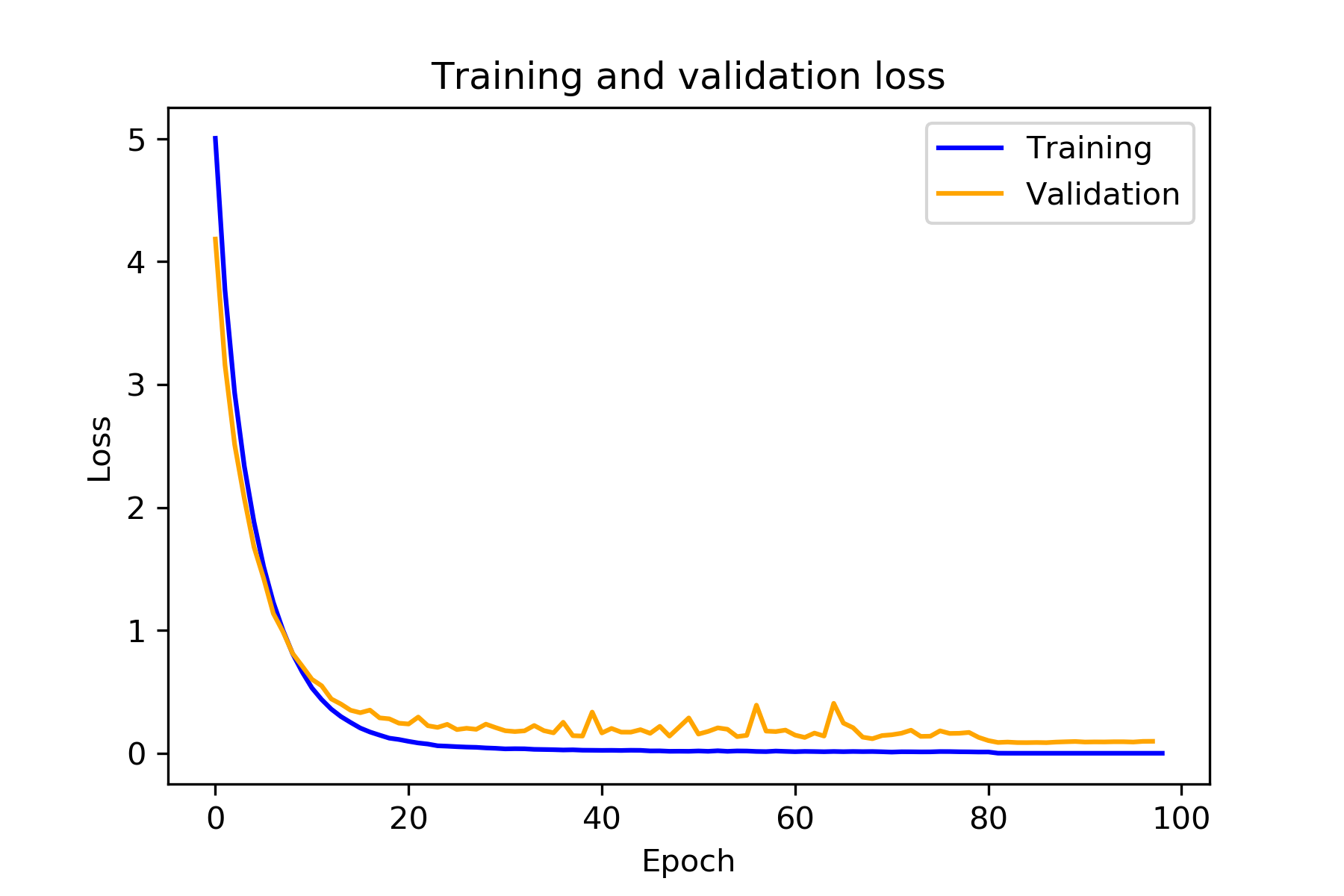}
		\caption{CNN+ FPM+ LSTM+ Attention}
		\label{fig:42_3}
	\end{subfigure}		
	\begin{subfigure}{0.32\textwidth}
		\includegraphics[width=\linewidth]{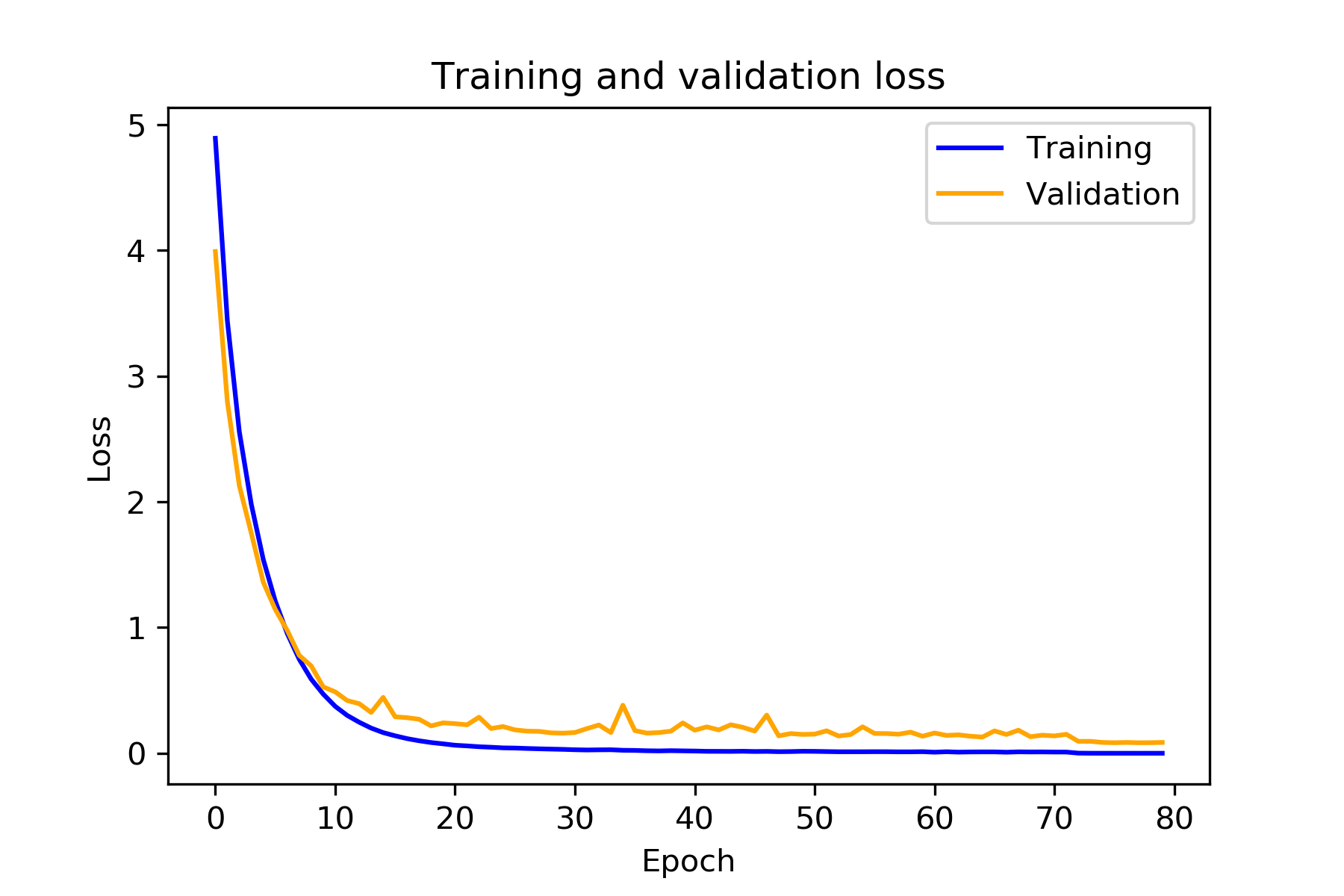}
		\caption{CNN + FPM + BLSTM + Attention}
		\label{fig:46_3}
	\end{subfigure}
\begin{subfigure}{0.32\textwidth}
	\includegraphics[width=\linewidth]{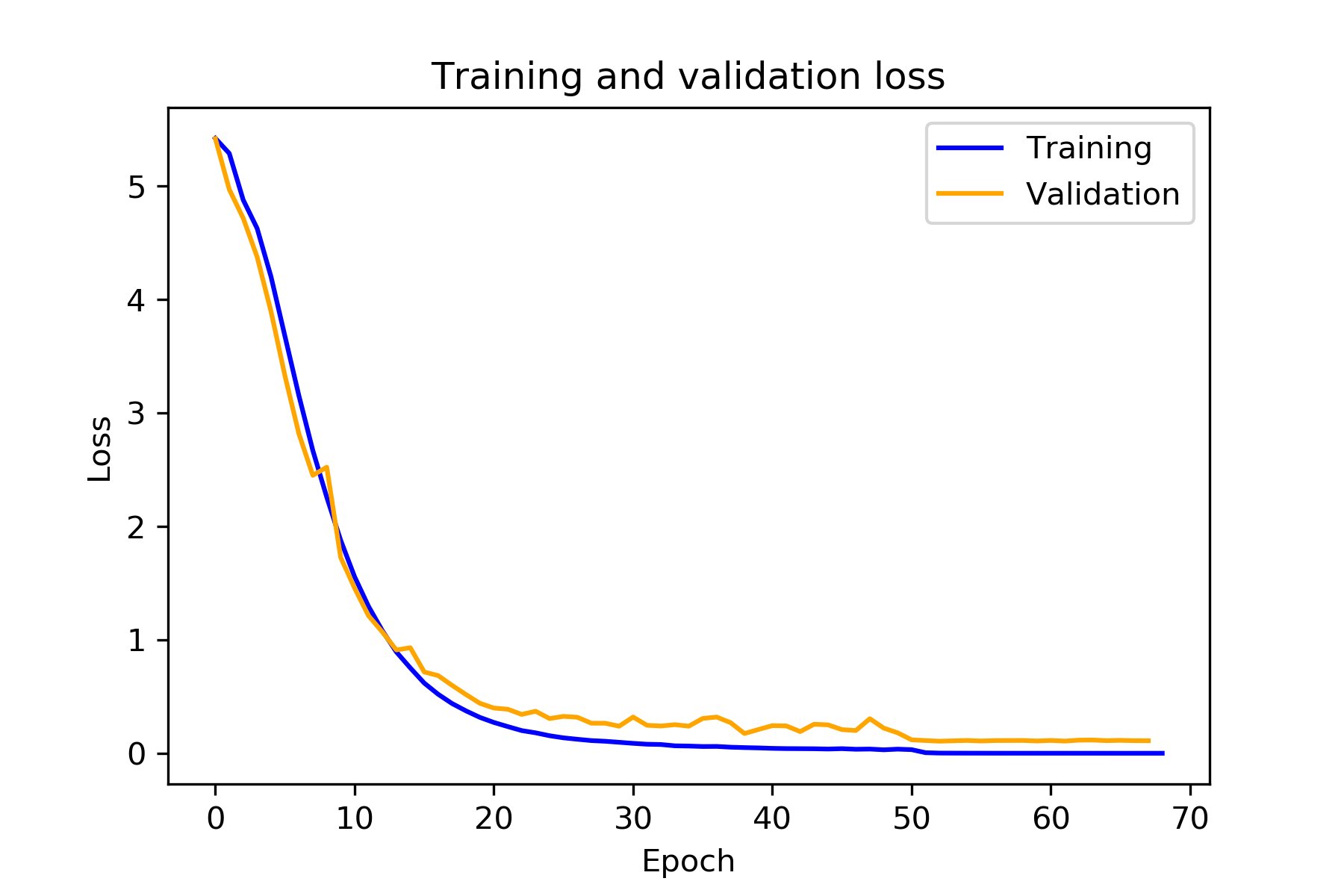}	
	\caption{CNN + LSTM}
	\label{fig:54_3}	
\end{subfigure}		
\begin{subfigure}{0.32\textwidth}
	\includegraphics[width=\linewidth]{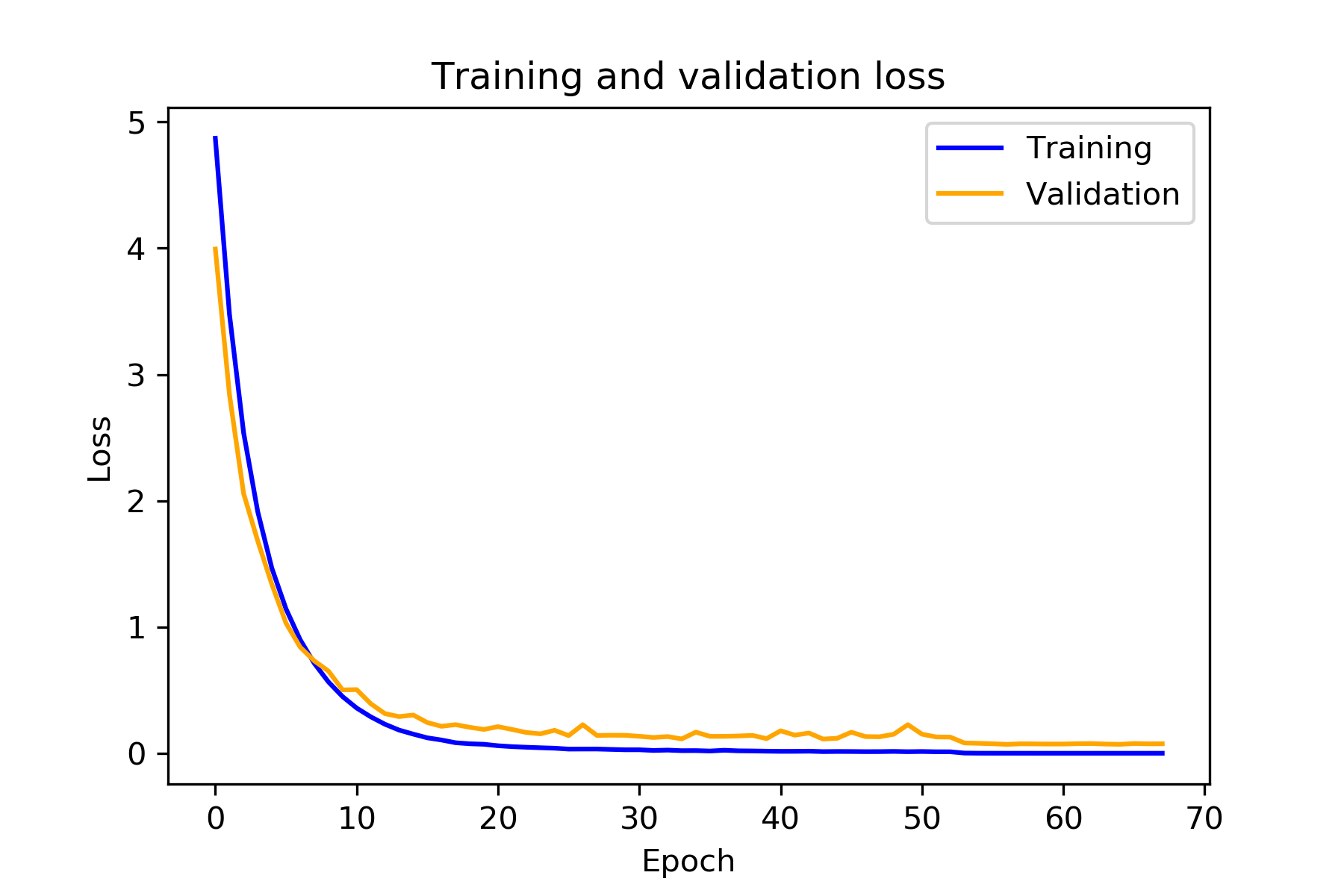}
	\caption{CNN + FPM + LSTM + Attention}
	\label{fig:52_3}
\end{subfigure}		
\begin{subfigure}{0.32\textwidth}
	\includegraphics[width=\linewidth]{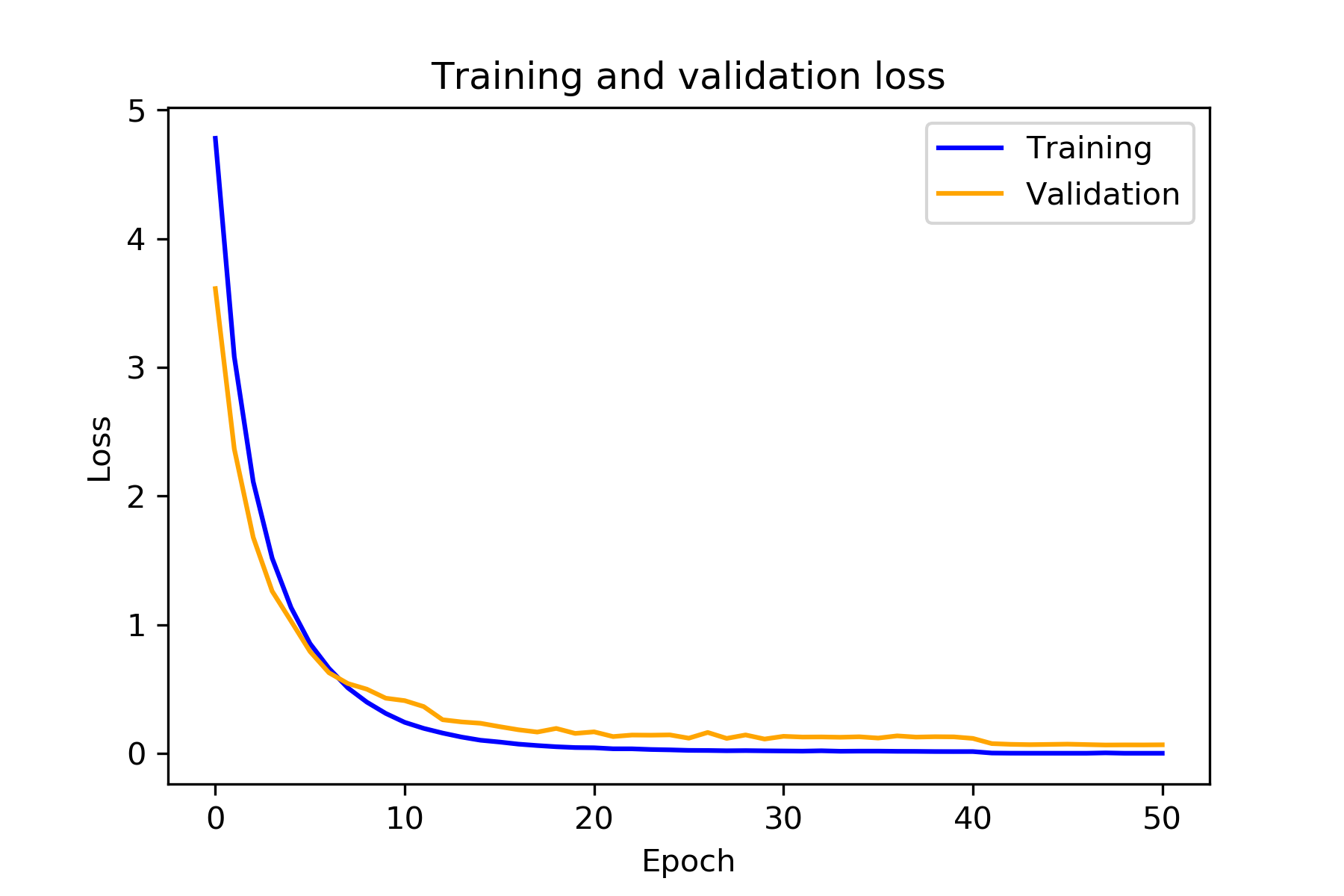}
	\caption{CNN + FPM + BLSTM + Attention}
	\label{fig:56_3}
\end{subfigure}
	\caption{Training and validation loss curves, (a, b, c) using only RGB data, (d, e, f) using RGB+Depth data on AUTSL dataset. }
	\label{fig9}       % Give a unique label
\end{figure}

\textit{\textbf{Model Training Times: }} We trained our models on NVIDIA Tesla V100. Table \ref{tab:times} shows the average training time of an epoch in our models. Training with RGB-D data takes almost two times more than training with RGB data only. While adding FPM to the network cause an increase in time, adding an attention mechanism do not increase the time as much. Moreover, adding an attention model enables the models to converge faster, as seen in Fig. \ref{fig9}. For example, in the case of using only RGB modality, training and validation losses get close to zero at around 40\textsuperscript{th} epoch with our vanilla model, i.e. \textit{CNN + LSTM}. On the other hand, attention-based models reach the same loss value at around 20\textsuperscript{th} epoch. Therefore, attention-based models converge faster than the other models in time. When we compare using single and multiple modalities, we observe that fusing RGB and depth data also reduces the total number of epochs during training. Moreover, validation losses are more stable in the case of using RGB-D data.

\begin{table}[t]
	\caption{Comparison of training time per epoch in hour on AUTSL.}
	\centering
	\label{tab:times}
	\begin{tabular}{ lcc }
		\hline
		\textbf{Model} & \textbf{RGB (hr)}  & \textbf{RGB+Depth (hr)} \\ \hline
		CNN + LSTM       & 2.2  & 4.3  \\ 
		CNN + FPM + LSTM     & -  & 5.8  \\ 
		CNN + LSTM + Attention       & 2.5 & 4.5 \\ 
		CNN + FPM + LSTM + Attention      & 3.5 & 6  \\ 
		CNN + FPM + BLSTM + Attention    & 3.7 & 6.3  \\ \\ \hline
		
	\end{tabular}
\end{table}

\textit{\textbf{Results on Signer Dependent Testing: }} We also conducted some experiments by randomly selecting training, validation, and test sets to show the model performances by training the models using all the signers, i.e. signer dependent model training. In this setting, all the signers in the test set are also included in the training and validation set; we randomly selected 72\% of all the videos for training, 13\% for validation and 15\% for testing. 

Since the model training takes too much time, we trained only two of our deep models; i.e. \textit{CNN + FPM + LSTM} and \textit{CNN + FPM + LSTM + Attention} models, end-to-end by using only RGB data, to obtain sample results to compare with corresponding user-independent models.  The architectures and the model parameters are kept exacly the same with our previous experiments; only the traning and test data selections are different. The results are shown in Table \ref{tab:autsl_random}. 
\begin{table} 
	\caption{Recognition rates (\%) of our models on AUTSL random test set.}
	\centering
	\begin{tabular}{l|ccc}
		\hline
		\multirow{3}{*}{\textbf{Method}} &
		
		\multicolumn{3}{c}{\textbf{Randomly Selected}}  \\ 
		& \multicolumn{3}{c}{\textbf{Train-Test Set}}  \\ 
		& {\textbf{top-1}} & {\textbf{top-3}} & {\textbf{top-5}} \\ \hline
		
		CNN + FPM + LSTM  & 94.07 & 98.79 & 99.36 \\
		CNN + FPM + LSTM + Attention   & 95.95 & 98.96 & 99.42 \\  \hline		
		
	\end{tabular}
	\label{tab:autsl_random}
\end{table}

These models work significantly better than their corresponding user-independent counterparts (Table \ref{tab:autsl_rgb}). We get 94.07\% with \textit{CNN + FPM + LSTM} model and 95.95\% accuracy with \textit{CNN + FPM + LSTM + Attention} in their top-1 accuracy. In this experiment, \textit{CNN + FPM + LSTM} performs already very high, hence the amount of performance increase with added attention model is small. Also, both models reached more than 99\% in their top-5 accuracies. These results show that our models work robustly when samples belonging to signers in the test set is viewed in the training set. When using the benchmark test data, which reflects the actual performances of the models in a realistic setting, the performances drop heavily in the models' top-1 accuracies. 

\subsection{Experiment Results on Montalbano}
\label{sec:results_montalbano}

Montalbano is a gesture dataset released by ChaLearn 2014 Looking at People Challenge which consists of 20 Italian gestures performed by 27 users. It contains 940 video sequences, each containing 10 to 20 gesture samples and around 14,000 samples in total (6,850 train, 3,454 validation, and 3,579 test samples). The videos are recorded with Microsoft Kinect in 640x480 pixel resolutions and four types of data are provided; RGB, depth, user segmentation, and skeleton.

\subsubsection{Preprocessing}
\label{sec:preprocessing}
We preprocess the videos in the Montalbano dataset similar to our preliminary work \cite{sincan2019isolated}. Since the problem we are dealing in this research is isolated sign language recognition, we created isolated sign samples from the Montalbano video sequences. We then cropped each frame from the upper body of the signers using the signer's shoulder center joint coordinates, using the skeleton data. After this operation, each frame size is fixed to 400x400 pixels. We kept the shoulder center point on the horizontal center line of the cropping square. In the vertical axis, the images are cropped by aligning the window to the upper part of the image. Furthermore, we also fixed the number of frames in all the videos to 40 frames as in \cite{sincan2019isolated}. 

\subsubsection{Training Details}
\label{sec:trainingDetails_montalbano}

We configured all our model experiments using the same hyperparameters as in the AUTSL experiments. However, in our experiments on the Montalbano dataset, we use the batch size as 16, since the videos have fixed number of frames. We set the initial learning rate as 1e-4 instead of 1e-5.

\subsubsection{Results}
We evaluate our best model, \textit{CNN + FPM + BLSTM + Attention}, on Montalbano dataset and we compare our results with the sate-of-art models that also work with Montalbano dataset in isolated recognition setting. Table \ref{tab:montalbano_results}  contains state-of-the-art model performances that use only RGB or RGB-D data. We achieved competitive results with the state-of-the-art models on this dataset without eager hyperparameter parameter tuning for this dataset; our model got 95.46\% accuracy using only RGB data and 96.11\% accuracy using RGB-D data.

\begin{table}[t]
	\caption {Comparison of the method performances on Montalbano dataset in isolated setting using RGB or RGB-D data.}
	\centering
	\label{tab:montalbano_results}
	\begin{tabular}{ llc }
		\hline
		\textbf{Model} & \textbf{Modality}  & \textbf{Recognition rate (\%)} \\ \hline
		\cite{molchanov2016online}       & RGB-D & 97.50 \\  
		\cite{pigou2018beyond}  & RGB-D & 97.23 \\  
		\textbf{Ours}    & \textbf{RGB-D} &  \textbf{96.11}  \\ 		
		\cite{neverova2015moddrop} & RGB-D & 95.06 \\ 		
		\cite{sincan2019isolated}       & RGB-D & 93.15 \\ 
		\cite{pigou2014sign}       & RGB-D & 91.70 \\  \hline
		\textbf{Ours}      & \textbf{RGB} &  \textbf{95.46} \\
		\cite{santos2020238}    & RGB  & 94.58  \\ 	\hline

	\end{tabular}
\end{table}

\section{Conclusion}
\label{sec:conclusion}
%\paragraph
In this paper, we present a new large-scale isolated Turkish Sign Language dataset that we named shortly as AUTSL. Our dataset provides various challenges compared to many other large-scale sign language datasets; to the best of our knowledge, it is the first large-scale public TSL dataset containing a variety of different backgrounds from indoor and outdoor settings that are performed with several different signers. In addition to the challenges provided with AUTSL, we aimed to perform user-independent classification of the signs in this research. We provide a benchmark training and test sets that we used in this research publicly available for the researchers. We also provide several deep learning-based models aiming to serve as baselines for future researches with this benchmark. 

We trained a series of models based on a vanilla \textit{CNN + LSTM} architecture. We incrementally integrated FPM and temporal attention to the vanilla model to improve the classification performances. All the models are trained with RGB-D data and RGB only data. Finally, we trained the best models of both modalities using BLSTM models replacing the LSTMs. The best results are obtained using RGB-D data using the \textit{CNN + FPM + BLSTM + Attention} architecture. In order to validate our baseline models, we also evaluated our best model architecture on Montalbano dataset and compared the performances with state-of-the-art approaches. Our models achieved competitive results with the state-of-the-art models using RGB and RGB-D data. We provided quantitative results using top-1, top-3 and top-5 classification accuracies of all the models on AUTSL dataset. The results reveal that some signs in our dataset are performed visually similarly and are misclassified by our models. Moreover, the challenges provided with variety of backgrounds that are gathered in unconstrained settings degrade the performance a lot. We provided sample visualizations of the spatially and temporally attended regions for some samples that support these claims. 

Providing these baseline models to the community, we also plan to work with AUTSL benchmark more to increase the classification performance in the future. We are planning to make more research to improve the spatial and temporal attention of our models to make them more robust to dynamic backgrounds. Moreover, we will focus more on better discriminative training of our models to increase the classification accuracy of similar signs in the future.

\section*{Acknowledgements}
This research is part of a project funded by TUBITAK (The Scientific and Technological Research Council of Turkey) under the grant number 217E022. The numerical calculations reported in this paper were partially performed at TUBITAK ULAKBIM, High Performance and Grid Computing Center (TRUBA resources). We would like to thank TUBITAK for the support. We also would like to thank to all our volunteer signers and to our TSL instructor Selda Demirci who contributed this research a lot voluntarily during data gathering.

\bibliographystyle{ieeetr}
\footnotesize \bibliography{AUTSLArxiv2020}

\end{document}